\PassOptionsToPackage{unicode}{hyperref}
\PassOptionsToPackage{hyphens}{url}
\PassOptionsToPackage{dvipsnames,svgnames,x11names}{xcolor}
\documentclass[
  11pt,
]{article}
\usepackage{amsmath,amssymb}
\usepackage{iftex}
\ifPDFTeX
  \usepackage[T1]{fontenc}
  \usepackage[utf8]{inputenc}
  \usepackage{textcomp} 
\else 
  \usepackage{unicode-math} 
  \defaultfontfeatures{Scale=MatchLowercase}
  \defaultfontfeatures[\rmfamily]{Ligatures=TeX,Scale=1}
\fi
\usepackage{lmodern}
\ifPDFTeX\else
\fi
\IfFileExists{upquote.sty}{\usepackage{upquote}}{}
\IfFileExists{microtype.sty}{
  \usepackage[]{microtype}
  \UseMicrotypeSet[protrusion]{basicmath} 
}{}
\makeatletter
\@ifundefined{KOMAClassName}{
  \IfFileExists{parskip.sty}{%
    \usepackage{parskip}
  }{
    \setlength{\parindent}{0pt}
    \setlength{\parskip}{6pt plus 2pt minus 1pt}}
}{
  \KOMAoptions{parskip=half}}
\makeatother
\usepackage{xcolor}
\usepackage[margin=1in]{geometry}
\usepackage{longtable,booktabs,array}
\usepackage{calc} 
\usepackage{etoolbox}
\makeatletter
\patchcmd\longtable{\par}{\if@noskipsec\mbox{}\fi\par}{}{}
\makeatother
\IfFileExists{footnotehyper.sty}{\usepackage{footnotehyper}}{\usepackage{footnote}}
\makesavenoteenv{longtable}
\providecommand{\tightlist}{%
  \setlength{\itemsep}{0pt}\setlength{\parskip}{0pt}}
\ifLuaTeX
  \usepackage{selnolig}  
\fi
\IfFileExists{bookmark.sty}{\usepackage{bookmark}}{\usepackage{hyperref}}
\IfFileExists{xurl.sty}{\usepackage{xurl}}{} 
\hypersetup{
  colorlinks=true,
  linkcolor={blue},
  filecolor={Maroon},
  citecolor={Blue},
  urlcolor={blue},
  pdfcreator={LaTeX via pandoc}}

\author{}
\date{}

\begin{document}

\hypertarget{memory-as-metabolism}{%
\section{Memory as Metabolism}\label{memory-as-metabolism}}

\hypertarget{a-design-for-companion-knowledge-systems}{%
\subsection{A Design for Companion Knowledge
Systems}\label{a-design-for-companion-knowledge-systems}}

\textbf{Stefan Miteski} CODE University Berlin
stefan.miteski@ext.code.berlin

April 2026 (v3.642)

\emph{This paper was developed with AI-assisted research and editing
support. The author takes full responsibility for all claims, framing,
citations, and conclusions.}

\begin{center}\rule{0.5\linewidth}{0.5pt}\end{center}

\hypertarget{abstract}{%
\subsection{Abstract}\label{abstract}}

Retrieval-Augmented Generation is still the dominant pattern for giving
LLMs persistent memory, but a visible cluster of personal wiki-style
memory architectures emerged in April 2026 --- design proposals from
Karpathy, MemPalace, and LLM Wiki v2 that compile knowledge into an
interlinked artifact for long-term use by a single user, instead of
retrieving from raw documents on every query. They sit alongside
production memory systems that the major labs have been shipping for
over a year, and an active academic lineage including MemGPT, Generative
Agents, Mem0, Zep, A-Mem, MemMachine, SleepGate, and Second Me. Within a
2026 landscape of emerging governance frameworks for agent context and
memory --- including Context Cartography ({[}45{]}) and MemOS ({[}24{]})
--- this paper proposes a companion-specific governance profile: a set
of normative obligations, a time-structured procedural rule, and
testable conformance invariants for the specific failure mode of
entrenchment under user-coupled drift in single-user knowledge wikis
built on the LLM wiki pattern {[}20{]}. This is an interpretive reading
of the field, not a settled diagnosis --- public docs may understate
internal theory, and the systems below were built for a range of
overlapping but distinct purposes. This paper contributes the design
language we found missing in our own reading.

The framework's three layers --- interaction/workflow (collect,
annotate, organize, revisit), representation/retrieval (storage format,
object types, retrieval index), and retention/governance (decay,
gravity, consolidation, audit) --- are explicitly scoped: this paper is
primarily a theory of the third, with implications for the second
through the CONTEXTUALIZE operation.

Start with how the wiki decides what to keep. Recency and access
frequency are not enough. An entry should also earn its place by being
structurally load-bearing for the rest of the wiki, and by the user
actually getting useful results when the system acts on it. Pure recency
retains popular noise; pure access frequency punishes quiet foundations;
pure utility creates a satisfaction-chasing echo chamber. You need all
three, weighted against each other, plus a defense against any one
signal dominating.

That defense is memory gravity. Some entries are referenced by many
others; removing them fragments the wiki and orphans downstream
knowledge. A naive pruner --- even a smart outcome-weighted one --- will
eventually delete the foundations while retaining what was popular last
week. Memory gravity protects entries whose removal would cascade, the
same way architectural gravity protects load-bearing decisions in
software systems. It is dependency centrality applied to memory
retention, and it is what keeps the wiki from optimizing itself into
fragility.

Then the hard question: what happens when new information arrives that
contradicts what is already in the wiki? Karpathy's lint operation
handles some of this reactively. Our proposal handles it the way humans
actually do --- not at ingestion, but later, during a dedicated
integration pass modeled on sleep consolidation. New entries land in a
raw buffer. A shallow filter rejects obvious garbage at ingestion but
does not make coherence decisions. The real coherence work runs on a
schedule (nightly, weekly, whatever fits), and it scores entries against
each other as well as against the active wiki. This matters because a
single contradictory entry scored alone gets quarantined, but three
mutually-supporting entries arriving in the same buffer window can
accumulate enough pressure to challenge a high-gravity dominant
interpretation. Minority positions get a structural path to becoming
majority positions when the evidence justifies it, instead of being
silently routed to quarantine one by one.

This connects directly to the Kuhn problem {[}21{]}. Wikis ossify. Over
time the dominant interpretation gets more protected, newer
contradicting evidence gets more easily dismissed, and what started as a
living knowledge base turns into a paradigm maintenance system ---
coherence preserved within the existing structure until anomalies
accumulate enough to force a shift. Normal science, not revolution. We
attack this with a periodic AUDIT operation that stress-tests the
highest-gravity entries --- temporarily suspends them, reruns queries
that used to access them, and measures whether query performance
actually degrades. If the entry turns out to be dead weight, its gravity
decays. If removing it improved query performance, it was actively
interfering and gets archived. AUDIT is the framework's main defense
against dogma, and we are honest that its sensitivity is an open
problem.

Underneath all of this is a design principle we think is missing from
current personal LLM memory discussions. Personal memory is a
\emph{companion system} --- its job is to serve one user over the long
haul, not to track objective truth. That changes the design target in
specific ways: the system should mirror its user on operational
dimensions (working vocabulary, load-bearing structure, continuity of
context) and compensate on epistemic failure modes (entrenchment,
suppression of contradicting evidence, the Kuhnian ossification
described above). Mirror-vs-compensate is not a slogan; it is a decision
rule with a temporal structure. The streaming path preserves usability
in the moment. The scheduled consolidation and audit operations
adjudicate revision.

One architectural commitment worth flagging up front: the wiki stays
outside the base model weights. This is deliberate. It preserves a
correction channel the framework does not implement and does not need to
--- when the base LLM gets updated by its lab (new factual priors, new
alignment training, new capabilities), the companion system inherits
those updates for free because swapping the model is a configuration
change, not a wiki operation. A user running this for five years is not
running the same reasoning engine at year five, even if the wiki looks
identical. Fold the wiki into weights and you lose this channel.

Finally, the scope question. Everything above concerns single-agent
systems. The interesting follow-on problems live at the federation level
--- family wikis, team wikis, department wikis, community wikis --- each
of which has different update dynamics (generational turnover,
onboarding, role handoffs, membership churn). Federation is a distinct
research direction, not a rescue for what single-agent systems cannot
do, and we name it as where the next layer of work belongs.

Honest scoping: the mechanisms are mostly borrowed from cognitive
architectures, recommender-system diversity, and graph centrality. The
contribution is how they are coordinated around a specific design rule
for a specific class of system. And honest safety: at the single-agent
level, the framework offers a partial story. It can resist entrenchment
and amplify genuine minority evidence. It does not eliminate the
reinforcement of user-held bad beliefs, and we do not pretend it does.
What we offer is an architecture that makes the gap visible, names three
correction channels that attack it on different timescales, and points
at the specific directions where the remaining work lives.

The safety story is explicitly partial: the framework offers structural
defenses on three timescales --- scheduled within-agent consolidation
cycles, cross-agent federation, and base-model evolution --- but does
not solve the reinforcement-of-bad-beliefs problem and does not claim
to.

\begin{center}\rule{0.5\linewidth}{0.5pt}\end{center}

\hypertarget{introduction}{%
\subsection{1. Introduction}\label{introduction}}

\hypertarget{three-claims-separately-defended}{%
\subsubsection{1.1 Three claims, separately
defended}\label{three-claims-separately-defended}}

Much of the difficulty in current LLM memory literature comes from
conflating three distinct claims. We separate them.

\textbf{The descriptive claim.} Personal LLM memory systems built around
incremental wiki compilation already exhibit user-coupled retention
dynamics that can accumulate into drift over time. Karpathy's LLM Wiki
{[}20{]} grows by integrating what the user finds worth integrating.
MemPalace {[}17{]} retains what the user's usage patterns reinforce. LLM
Wiki v2 {[}11{]} decays what the user does not return to. None of these
systems is a neutral knowledge infrastructure. Each is position-ful and
personalized, even when its documentation does not say so.

\textbf{The taxonomic claim.} This drift is sufficient grounds to
specify the retention-governance obligations that follow from treating
personal single-user LLM memory as a distinct design class --- a class
that prior work has named without governing. Human-centered computing,
user modeling, adaptive interfaces, and personal knowledge management
have long drawn similar distinctions between systems that serve a
specific user's goals and systems that serve a general information need.
We are not the first to draw this line. MemoryBank (arXiv:2305.10250)
already uses ``long-term AI Companion scenario'' as a primary capability
descriptor. Second Me (arXiv:2503.08102) explicitly frames single-user
memory as a ``memory offload system'' serving one user. LongMemEval
(arXiv:2410.10813) ties long-term memory evaluation to contexts
including ``psychological counseling or secretarial duties.'' Recent
surveys (spanning 2024--2026) treat ``personalized'' as an established
memory category ({[}13{]}; {[}44{]}; {[}51{]}). What is missing is not
the word \emph{companion} but a normative specification of what
retention-governance obligations follow from treating a system as one
--- what it must mirror, what it must compensate for, what separability
requires, and why those obligations make the class designable in ways
that ``personalized'' or ``companion-scenario'' alone do not. That
specification is what this paper provides.

\textbf{The normative claim.} Given the descriptive fact and the
taxonomic distinction, the retention policy for companion memory should
follow a specific design principle we introduce in Section 1.2 ---
mirror on operational dimensions, compensate on epistemic failure modes.
This is where the contribution lives. The descriptive claim is
observation; the taxonomic claim is a classification move others have
made in adjacent domains; the normative claim is the design rule we
propose and defend.

Each claim stands alone. The taxonomic claim does not depend on agreeing
with the normative one. The normative claim does not assume the
descriptive claim is settled --- it gives a design rule that remains
useful even if reasonable people disagree about whether current systems
already drift.

\hypertarget{mirror-where-mirroring-serves-utility.-compensate-where-mirroring-damages-it.}{%
\subsubsection{1.2 Mirror where mirroring serves utility. Compensate
where mirroring damages
it.}\label{mirror-where-mirroring-serves-utility.-compensate-where-mirroring-damages-it.}}

A companion system should mirror its user on some dimensions and
compensate for its user on others. The selection rule is instrumental,
not philosophical.

A companion system mirrors its user on \emph{operational} dimensions:
the working context the user is currently reasoning within, the
load-bearing structure the user depends on for coherent thought, the
continuity of self-reference that lets the user pick up where they left
off, the vocabulary and framing the user has developed over time. On
these dimensions, alignment is the design goal and deviation is the
failure mode. A companion that refused to inherit its user's vocabulary
would not be usable.

A companion system compensates for its user on \emph{epistemic failure}
dimensions: entrenchment of demonstrably false high-gravity entries,
suppression of evidence contradicting settled beliefs, convergence
toward monoculture under repeated use. On these dimensions, alignment is
the failure mode and deviation is the design goal. A companion that
inherited its user's every confirmation bias would be harmful.

The mirror-vs-compensate vocabulary itself is not new. Qian et
al.~(arXiv:2510.01924) uses it explicitly in ``To Mask or to Mirror''
--- empirically observing that some models mirror human biases while
others mask and compensate for them at inference time, and naming the
tension in terms close to this paper's own. The sycophancy literature
frames over-mirroring as a failure mode requiring procedural mitigation.
The offline consolidation pattern that implements the compensate side is
also established: LightMem (arXiv:2510.18866) explicitly frames its
design as ``sleep-time computation that decouples consolidation from
online inference,'' and SleepGate (arXiv:2603.14517) proposes periodic
sleep micro-cycles for KV-cache consolidation. What neither the
vocabulary nor the individual mechanisms provide is a
\emph{time-structured procedural conflict rule} for resolving the
mirror-vs-compensate tension in a personal companion-memory substrate
--- specifically, a decision procedure governing what gets buffered
versus quarantined versus audited, across which timescale, and with what
decision consequences at each stage. The contribution is the TRIAGE →
CONSOLIDATE → AUDIT execution model as a binding: not the discovery of
the tension, and not the individual operations, but the procedural rule
that decides how and when each operation applies to the
mirror-vs-compensate conflict in a companion wiki.

The analogy is corrective rather than substitutive. A cane mirrors the
user's gait --- it does not try to walk differently. A cane does not
mirror a limp --- it compensates for it. Glasses mirror the user's
visual field and compensate for its distortion. Companion memory should
mirror operational continuity and compensate for epistemic entrenchment.
The framework's five operations implement this split. TRIAGE, DECAY, and
memory gravity are mirror mechanisms. CONSOLIDATE, AUDIT, and
CONTEXTUALIZE are compensate mechanisms --- CONTEXTUALIZE in a more
specific sense, addressed in Section 5.4 below: it compensates for the
assumption that external sources have a single canonical compression by
fitting them to the user's working context depth at consolidation time.

The tension between mirror and compensate is not a bug to resolve. It is
the design principle. Any companion memory framework that claims to only
mirror is unsafe; any framework that claims to only compensate is not a
companion.

When mirror and compensate point in opposite directions, the framework
defaults to preserving operational continuity in the streaming path and
routing the conflict to scheduled compensate operations. A single
contradiction should not overwrite a high-gravity entry in real time ---
that would destroy the continuity the companion is supposed to provide.
But coherence concerns should not be allowed to suppress accumulated
counterevidence indefinitely either. TRIAGE preserves usability in the
moment. CONSOLIDATE and AUDIT decide whether revision has earned
structural change. The rule is procedural rather than algorithmic:
mirror by default under time pressure, compensate during scheduled
integration windows, and treat AUDIT as the tiebreaker when a
gravity-protected entry is implicated in repeated bad outcomes across
multiple cycles. (See §5.0 for the conflict routing matrix that
instantiates this rule case-by-case.)

\hypertarget{the-circularity-is-the-thesis}{%
\subsubsection{1.3 The circularity is the
thesis}\label{the-circularity-is-the-thesis}}

A frequent objection to coherence-based memory policies is that they are
self-sealing: coherence is measured against the current wiki, which is
the product of past coherence decisions. Under a truth-tracking framing,
this would be a fatal flaw. Under the companion framing, it is what
having a stable self looks like rather than dissociating every time new
information arrives. We accept the circularity on the mirror side and
build the framework's compensate side --- batched consolidation and
audit --- specifically to resist its failure modes without pretending to
escape it.

\hypertarget{contributions}{%
\subsubsection{1.4 Contributions}\label{contributions}}

\begin{enumerate}
\def\labelenumi{\arabic{enumi}.}
\tightlist
\item
  A triple-tracked framing of personal LLM memory --- descriptive,
  taxonomic, normative --- that separates observation, classification,
  and design rule rather than collapsing them.
\item
  The mirror-vs-compensate design principle as an instrumental selection
  rule for which user properties a companion system should inherit,
  operationalized as a time-structured procedural conflict rule across
  streaming, consolidation, and audit timescales.
\item
  A five-operation retention policy (TRIAGE, CONTEXTUALIZE, DECAY,
  CONSOLIDATE, AUDIT) built around a raw buffer and a batched
  consolidation cycle modeled on sleep function.
\item
  Two supporting mechanisms: memory gravity (load-bearing protection for
  operational continuity) and minority-hypothesis retention (variance
  against monoculture collapse).
\item
  Four predictions with operational proxies: coherence stability,
  fragility resistance, monoculture resistance, effective
  minority-hypothesis influence --- where influence is defined as
  measurable change in downstream outputs, not mere storage or
  surfacing.
\item
  An honest safety story with three correction channels --- within-agent
  consolidation, cross-agent federation across named unit types, and
  base model evolution preserved by architectural separability --- and
  explicit acknowledgment of what the framework does not solve.
\end{enumerate}

We do not present implementation results. This is a vision paper
proposing a normative governance profile for a specific system class.

\begin{center}\rule{0.5\linewidth}{0.5pt}\end{center}

\hypertarget{background-and-related-work}{%
\subsection{2. Background and Related
Work}\label{background-and-related-work}}

\hypertarget{llm-memory-systems}{%
\subsubsection{2.1 LLM Memory Systems}\label{llm-memory-systems}}

\emph{A note on sources: this paper draws from peer-reviewed
publications, arXiv preprints, practitioner-published design documents
(including GitHub gists), and community-reported analyses of production
systems. These carry different evidential weight. Where a claim rests
primarily on community reporting rather than official documentation, the
text marks this explicitly. The community-reported characterizations of
production systems --- particularly the Auto Dream mechanism attributed
to Anthropic --- are included as motivating context for the design
framework, not as settled empirical claims.}

The literature map below is interpretive and necessarily incomplete; it
maps the design space as the author reads it, not as a systematic review
would establish it.

Retrieval-Augmented Generation {[}23{]} treats every query as a fresh
retrieval problem --- knowledge is indexed but never consolidated. Three
open projects proposed alternatives in April 2026: Karpathy's LLM Wiki
{[}20{]} addresses the stateless-LLM problem by proposing that the model
should incrementally compile a persistent interlinked knowledge base as
it reads new sources; MemPalace (Jovovich \& Sigman, 2026) adds a
hierarchical spatial retrieval architecture reporting 96.6\% R@5 on
LongMemEval; LLM Wiki v2 {[}11{]} adds Ebbinghaus-inspired time decay
and consolidation tiers. Obsidian is Karpathy's rendering surface; the
graph structure follows from interlinking rather than being the design
goal.

\textbf{Karpathy's LLM Wiki is the clearest open substrate this paper
directly governs.} Karpathy identifies three architectural layers ---
raw sources (immutable input), the compiled wiki (LLM-maintained
markdown files), and the schema (the configuration document telling the
LLM how the wiki is structured and what workflows to follow). He notes
that ``the wiki is just a git repo of markdown files'' and explicitly
leaves the schema ``intentionally abstract,'' designed to be co-evolved
between the user and the LLM for each domain. The companion governance
profile proposed in this paper is a governed instantiation of that
schema: normative obligations, vitality mechanics, and structural audit
cycles that Karpathy left to each user to instantiate. His pattern
identifies the substrate; this paper specifies the obligation-level
rules that should govern it for companion memory specifically. The
integration with existing systems is additive: Karpathy's LLM Wiki
provides the incremental compilation pattern; the companion layer adds a
raw buffer tier, a scheduled consolidation cycle, and metadata
structures for gravity and cohesion tracking.

\textbf{LLM Wiki v2} {[}11{]} extends Karpathy's base pattern with
explicit lifecycle governance proposals: Ebbinghaus-inspired retention
curves, multi-tier consolidation (working/episodic/semantic/procedural),
schedule-driven maintenance, and audit trail recommendations. This is
the closest practitioner-level prior art to the governance profile this
paper proposes, and the delta must be stated explicitly. LLM Wiki v2
proposes these mechanisms informally as implementation patterns without
normative obligations --- it does not specify what a companion wiki MUST
do, what it MUST NOT do, or what failure looks like for the specific
entrenchment failure mode. This paper adds three things LLM Wiki v2 does
not provide: (1) explicit entrenchment stress-testing via
AUDIT-by-suspension tied to utility traces; (2) explicit minority
retention across cycles against centrality-protected incumbents, with
multi-cycle buffer pressure as the integration mechanism; and (3)
companion-specific normative obligations that distinguish what should be
mirrored from what should be compensated, and why.

These projects are not the whole state of the art. Much of the relevant
frontier lives in production at the major LLM labs and in an active
academic literature, and both have been working on LLM memory for over a
year with substantial deployment data. A proper engagement with prior
work has to credit them honestly. \emph{The reading of these systems
below is interpretive rather than exhaustive; the point is to locate
common architectural tendencies, not to provide definitive product
analysis. Some of these systems target overlapping but distinct
problems, and grouping them should not be read as a claim that they are
all solving the same task.}

\textbf{Anthropic (Claude Code).} CLAUDE.md hierarchical memory with
scope-based instruction loading; per Anthropic's documentation, more
specific contexts take precedence over broader ones, with project-level
instructions taking priority over user-level instructions in conflict
resolution. Auto Memory for session-level capture. A between-session
consolidation mechanism --- referred to as \textbf{``Auto Dream''} in
community documentation and third-party analysis, though not named as
such in official Anthropic docs --- is reported to perform contradiction
resolution, date normalization, stale-entry pruning, and overlap
merging, with the sleep/REM framing representing community
interpretation rather than Anthropic's own characterization. Memory Tool
API with view/create/str\_replace/insert/delete/rename primitives
(officially documented). If the community-reported consolidation
behavior is accurate, it is the closest shipping parallel to this
paper's CONSOLIDATE operation. Anthropic ships mechanisms without
publishing design principles.

\textbf{OpenAI (ChatGPT).} Memory live since 2024, with Saved Memories
(explicit) and Reference Chat History (implicit) mechanisms. Community
analysis has proposed a four-layer architecture that is not RAG --- no
vector database, no embedding similarity search (Khemani, 2025,
community reverse-engineering; not official OpenAI documentation).
User-visible inspection, edit, delete. ChatGPT Memory is configured
through ChatGPT settings and operates at the product layer; the OpenAI
Responses API separately supports stateful conversation interactions,
which are architecturally distinct from the ChatGPT memory features.

\textbf{Google (Gemini / NotebookLM).} Gemini 3 Pro has a 1M token
context window. Google's strategy is effectively context-window
expansion as a partial substitute for sophisticated memory architecture
--- if you can fit everything in context, the forgetting problem is
deferred rather than solved. This is architecturally opposite to the
framework we propose, and worth engaging as an alternative design
philosophy.

\textbf{DeepSeek-OCR} (Wei, Sun, Li, arXiv 2510.18234, October 2025). An
open-source proposal to implement forgetting via progressive visual
compression. Long contexts are rendered as images, and older content is
progressively resized to smaller, blurrier images over time. 10x
compression holds at \textasciitilde97\% OCR accuracy, 20x holds at
\textasciitilde60\%. The paper frames this as a ``memory forgetting
mechanism in LLMs'' and draws the biological analogy explicitly ---
recent memories stay sharp, old memories fade. DeepSeek-OCR operates at
a different architectural layer than our framework: it proposes a
\emph{mechanism} for graceful forgetting, we propose a \emph{design
language} for retention decisions such mechanisms implement. The two
compose rather than compete. Subsequent work has contested the specific
mechanism --- Context Cascade Compression (C3, arXiv 2511.15244)
achieves 93\% at 40x via cascading two LLMs, and an adversarial critique
(arXiv 2512.03643) argues simple mean pooling outperforms the optical
approach at matched budgets --- which demonstrates that the mechanism
space for LLM forgetting is actively contested.

\textbf{Academic lineage --- Generative Agents and MemGPT.} Park et
al.~(arXiv 2304.03442, UIST 2023) introduced the ``memory stream'' --- a
comprehensive raw record of all agent experiences --- combined with
periodic ``reflection'' synthesis into higher-level abstractions. This
is a clear structural precursor to the pattern this paper builds on: a
TRIAGE-filtered raw capture feeding a buffer that a scheduled
CONSOLIDATE operation processes into a longer-term store. Generative
Agents is highly cited and must be acknowledged directly: the
buffer-plus-reflection pattern is prior art at the mechanism level. What
the companion framework adds is a normative design rule governing
\emph{which retention decisions are right} within that pattern, and a
named correction channel through architectural separability. Packer et
al.~(MemGPT, arXiv 2310.08560, 2023) then formalized LLM memory as
\emph{virtual context management} borrowed from operating-system virtual
memory paging: the LLM manages what sits in its own context window via
function calls, moving data between main context and external archival
storage. MemGPT explicitly names ``virtual companions or personalized
assistants'' as the use case where context budgets run out quickly,
which is the same class of system this paper theorizes. MemGPT's
architecture is silent on \emph{which retention decisions are right} ---
it provides the machinery for moving data between tiers without a design
language for deciding what belongs in which tier.

\textbf{Second Me} (arXiv 2503.08102). An intelligent, persistent memory
offload system that retains user-specific knowledge across contexts for
a single user. In this paper's taxonomy --- not a claim Second Me makes
about itself --- it is among the closest existing system-level neighbors
to the companion-memory design class this paper names. The distinction
is normative rather than architectural: Second Me describes the class
behavior and builds a working system; this paper provides the evaluation
target and the retention-policy obligations that should govern such
systems. Second Me does not specify what the system must mirror, what it
must compensate for, or what separability requires --- the normative
specification is what is absent, and that absence is what this paper
fills.

\textbf{Successor systems and the sleep-consolidation pattern.}
SleepGate (arXiv 2603.14517) proposes ``sleep micro-cycles'' over the
KV-cache with a forgetting gate and consolidation module, triggered
periodically. This complements LightMem (arXiv:2510.18866), which frames
its offline consolidation as ``sleep-time computation that decouples
consolidation from online inference.'' Taken together, SleepGate,
LightMem, and the community-reported Auto Dream mechanism suggest that
the sleep-consolidation pattern is emerging independently across
multiple architectural layers --- KV-cache, external wiki, and
companion-memory retention policy. The design vocabulary for why it
belongs specifically in companion-memory systems is what is missing.
Mem0 (Chhikara et al., arXiv 2504.19413, April 2025) is the strongest
documented production-paper in this lineage, proposing dynamic
extraction, consolidation, and retrieval of salient information from
ongoing conversations with a graph variant for relational structures,
and reporting a 26\% improvement over OpenAI on LLM-as-a-Judge on the
LOCOMO benchmark plus 91\% lower p95 latency and 90\%+ token cost
savings; Zep uses a temporal knowledge graph with time-aware validity
windows (Zep / Graphiti, arXiv 2501.13956); A-Mem (arXiv 2502.12110)
reports an 85-93\% token reduction versus MemGPT via interconnected
atomic notes; MemMachine (arXiv 2604.04853, March 2026) explicitly
grounds its design in Tulving's episodic/semantic distinction and
critiques the MemGPT lineage for ``accuracy concerns from probabilistic
extraction and compounding error over time.'' MemMachine's critique is
essentially the fragility-under-drift problem this paper's CONSOLIDATE
operation addresses --- the academic community is converging on the same
failure mode we are, at roughly the same time, and trying to solve it at
the mechanism layer.

\textbf{MemOS} ({[}24{]}) is the closest prior art to the
retention/governance layer this paper proposes and must be engaged
directly. MemOS explicitly names ``Memory Governance'' as a design stage
with access control, versioning, and provenance auditing. It uses an
L0/L1/L2 tiering structure --- raw (L0), structured (L1), and
internalized preferences (L2) --- that maps structurally onto the cold
memory / raw buffer / active wiki architecture proposed here. MemOS
frames governance as a safety foundation, which aligns with the
companion framework's treatment of AUDIT and minority-hypothesis
retention as safety mechanisms rather than optional optimizations. The
distinction is normative rather than architectural: MemOS describes
governance as an engineering architecture; this paper specifies the
obligation-level rules that should govern it for companion systems
specifically --- what the system must mirror, what it must compensate
for, and why those obligations make the governance rules non-optional
for this class. MemOS does not specify what a companion wiki must retain
as a function of epistemic drift, what it must compensate for as a
function of entrenchment, or why separability is a safety commitment
with a named rationale. The normative specification is what the
companion framework adds to the governance architecture MemOS names.

\textbf{The harness engineering review} ({[}53{]}) explicitly recommends
cross-model transfer tests for memory systems and argues that
reliability gains come from changing the environment around the base
model rather than the model itself --- directly encroaching on the
separability contribution. The companion framework's response is in
Section 8.3: the paper's separability claim is not that external memory
is more reliable (the harness review's framing) but that separability
specifically preserves the base-model evolution correction channel
against user-coupled epistemic entrenchment. That rationale is narrower
and more specific than the harness review's argument, and it remains
intact.

\textbf{Context Cartography} ({[}45{]}, March 2026) is the closest
competitor at the governance-layer level and must be engaged directly.
It proposes a formal framework for the deliberate governance of
contextual space with seven cartographic operators, explicit state
transitions, failure mode classification, and a diagnostic benchmark
designed for operator ablation --- parts of which are mechanically
verified in Lean 4. This occupies the governance-layer slot this paper
also targets, and the distinction must be stated precisely. Context
Cartography governs contextual space as a general problem: zones,
operators, generalization across agent types. This paper governs a
specific failure mode in a specific system class: entrenchment under
user-coupled drift in single-user companion wikis. The contribution is
not the existence of a governance layer --- Context Cartography and
MemOS both provide governance layers --- but a companion-specific
normative profile: what a single-user wiki MUST do when faced with the
specific failure mode of coherence-preserving drift that gradually
protects dominant interpretations against legitimate revision. The
evaluation target also differs: Context Cartography evaluates via
operator ablation; this paper's sharpest prediction targets multi-cycle
buffer pressure accumulation under centrality-protected entrenchment,
which no existing operator ablation scheme captures.

\textbf{What all of this means for this paper's contribution.}
Mechanism-level parallels are substantial and honest engagement requires
acknowledging them. At least six mechanisms in our framework have direct
parallels in shipped or published work: consequence-weighted retention
(OpenAI, Anthropic), batched consolidation (Anthropic,
community-reported; Generative Agents, 2023), hierarchical memory with
scope-based precedence tiers (Anthropic CLAUDE.md), contradiction
resolution during integration (Anthropic, community-reported),
stale-entry pruning (Anthropic, community-reported), user-managed
forgetting (OpenAI, Anthropic). The framework we propose is not a claim
to have invented the mechanisms. It is a claim to provide design
vocabulary the field has not assembled in this form --- a named system
class with \emph{normative obligations}, a principled retention policy
for that class, and an honest scope of what such systems can and cannot
do. Section 8.2 engages this more fully.

\textbf{The intent taxonomy gap.} The field now has at least six major
survey papers (2024--2026) and they are genuine contributions. The AI
Hippocampus (Jia et al., arXiv 2601.09113, TMLR 2025) organizes LLM
memory into implicit/explicit/agentic paradigms by representational
substrate. {[}13{]} uses forms/functions/dynamics. {[}44{]} uses
object/form/time. {[}51{]} uses sources/forms/operations. None of these
is a companion-memory taxonomy. All are mechanism taxonomies --- they
distinguish systems by \emph{how} they work, not by \emph{what they are
for}. What existing taxonomies do not provide is an \emph{intent
taxonomy}: an account of which evaluation target and which design
obligations distinguish one system class from another. A companion
memory system and a memory platform may share identical architectural
mechanisms but differ in what success looks like and what
retention-policy obligations follow. The AI Hippocampus's
implicit/explicit/agentic axis is orthogonal to this paper's
companion-memory class; citing it here is not a concession but a
demonstration that the field's best surveys are organized by substrate
while the design-class distinction this paper draws operates on a
different axis entirely. The intent taxonomy is what this paper names.

\hypertarget{user-aligned-system-design-as-prior-art}{%
\subsubsection{2.2 User-Aligned System Design as Prior
Art}\label{user-aligned-system-design-as-prior-art}}

The distinction between systems built to serve a specific user and
systems built to serve general information needs is not new to this
paper. Human-centered computing has developed it across decades of
adaptive interface, user modeling, and personalization research.
Personal knowledge management (PKM) treats individual knowledge bases as
artifacts whose value is measured by personal utility rather than
objective completeness. Communities of practice literature addresses how
small groups maintain shared knowledge that serves the group without
claiming general validity. We inherit the distinction from these
traditions and apply it to LLM memory, where it has not been drawn
cleanly in current discussions.

Wikipedia's governance architecture is a structurally instructive
parallel for the CONSOLIDATE and AUDIT operations, worth noting before
those operations are specified in Section 5. Wikipedia maintains
coherence at civilizational scale through a multi-agent governance
model: a large population of editors contribute changes to a shared
knowledge base, while a structured system of talk pages, revision
history, and flagged revisions adjudicates disputes and contested
claims. The talk page process is structurally analogous to CONSOLIDATE
--- contested claims are scored against existing content, minority
positions are preserved and visible rather than silently overwritten,
and integration decisions emerge through a deliberation process. The
featured article and good article review processes are structurally
analogous to AUDIT --- high-prominence articles are periodically
stress-tested by reviewers who temporarily impose additional scrutiny to
confirm their quality remains load-bearing. Two key differences mark the
companion framework as distinct: Wikipedia's governance is multi-agent
and human-operated, while the companion framework's governance is
single-agent and automated. These differences are design features, not
limitations --- the companion framework achieves faster consolidation
cycles precisely because it does not wait for human editorial consensus,
and its single-user scope makes the multi-agent coordination problem
unnecessary. The Wikipedia parallel is useful as a template for the
governance logic, not as a claim that the two systems are equivalent in
scope or mechanism.

A second important lineage comes from formal belief revision and truth
maintenance systems in knowledge representation. In the AGM framework
and related work on epistemic entrenchment (Gärdenfors \& Makinson,
1988), revision policies rely on an ordering over beliefs that
determines retraction priority when new information creates
inconsistency. Memory gravity functions as a pragmatic, graph-based
proxy for this kind of entrenchment ordering: it protects structurally
load-bearing entries based on downstream fragmentation cost rather than
purely logical postulates --- and the key difference is that gravity is
prospective (what would break if this entry were removed now) rather
than retrospective (what has historically referenced it).

Complementing this, Truth Maintenance Systems {[}6{]} maintain explicit
dependency networks and justifications, allowing selective retraction
while preserving consistent alternatives. The minority-hypothesis
retention and branch mechanism proposed here draws structural
inspiration from TMS-style dependency-directed revision, but adapts it
to an LLM-compiled wiki substrate: alternatives are kept alive across
scheduled consolidation cycles rather than under immediate logical
revision, and promotion decisions are driven by accumulated pragmatic
utility and multi-cycle buffer pressure rather than symbolic consistency
alone. These traditions therefore inform the governance profile
developed in Section 5, while the companion-specific normative
obligations --- mirror operational continuity while compensating
epistemic failure under user-coupled drift --- remain the distinguishing
contribution of this framework.

\hypertarget{individual-memory-models-and-sleep-consolidation}{%
\subsubsection{2.3 Individual Memory Models and Sleep
Consolidation}\label{individual-memory-models-and-sleep-consolidation}}

ACT-R's base-level learning {[}1{]} models memory activation as a
function of use history. Ebbinghaus's forgetting curve {[}7{]} grounds
spaced repetition. Tulving's episodic/semantic distinction {[}40{]} maps
onto the raw buffer and active wiki tiers in our framework.

More directly relevant to Section 5 is the cognitive neuroscience
literature on sleep consolidation. Tononi's synaptic homeostasis
hypothesis and McClelland's complementary learning systems both describe
the same basic pattern: episodic experience accumulates in a
fast-learning buffer during waking hours, and the deep integration work
--- coherence checking, contradiction resolution, transfer to long-term
stable structure --- happens offline during sleep. This is more than a
decorative analogy. It provides an architectural template for how
bounded agents can separate rapid capture from slower
coherence-preserving integration, and our framework implements the same
pattern at the architectural level without claiming the
neuroscience-to-system mapping is mechanistically exact.

A third precedent worth flagging --- though we treat it as future-work
direction rather than as an established building block --- is the
tiered-coherence pattern found in mature legal systems. European civil
law systems maintain coherence at civilization scale by layering
knowledge with different update frequencies and procedural protections:
constitutional principles that change across generations, primary
legislation that changes across decades, secondary regulation that
changes across years, and case law that updates continuously. This
shares structural features with the design problem the companion memory
framework addresses, though the legal solution rests on legitimacy,
authority, and institutional process --- not just coherence management.
The framework's gravity model implicitly points in a similar direction,
and a natural extension worth exploring would formalize gravity as a
discrete tier structure with different procedural protections per tier.
One honest caveat: the legal-system precedent does not promise
determinism. The same text --- for example Council Regulation (EEC)
3720/85 --- has produced materially different implementations across
member states, because legal application has interpretive latitude the
text does not eliminate. The same is true for companion memory wiki
content read by an LLM. Fixed text yields bounded variation in
downstream behavior, not a single deterministic output. The framework
should be read as constraining the range of LLM interpretation rather
than collapsing it; this is a structural property of the architecture,
not a bug to fix. A proper engagement with legal scholarship on tiered
coherence --- Hart's primary/secondary rules, Luhmann's
systems-theoretic treatment of law, comparative constitutional design
--- is reserved for future work.

\hypertarget{pragmatism-and-the-re-entry-of-truth-through-consequences}{%
\subsubsection{2.4 Pragmatism and the Re-entry of Truth Through
Consequences}\label{pragmatism-and-the-re-entry-of-truth-through-consequences}}

Consequence-weighted retention has serious philosophical precedent. The
pragmatist tradition --- Peirce, James, Dewey --- treats beliefs as
tools for action whose value is measured by practical consequences
rather than correspondence with an external reality the agent cannot
directly access.

This does not exempt companion systems from truth concerns. It reroutes
them. Under pragmatism, false beliefs become problematic when they
produce failed actions --- damaged plans, broken relationships, health
harms, safety failures. Truth objections do not disappear; they come
back through consequences. The framework's utility signal is the channel
for this re-entry. A retained entry that consistently produces bad
outcomes loses its retention whether or not anyone has labeled it
\emph{false}. Pragmatism lets the framework sidestep correspondence as a
design target while keeping consequence tracking as a correction
mechanism.

The apparent tension between ``not a truth-tracker'' and ``uses
consequence as a retention signal'' resolves under pragmatism. The
framework does not measure correspondence with external reality --- it
has no truth oracle. What it measures is pragmatic fitness: did acting
on this entry produce outcomes the user judged useful? Did its presence
improve or degrade subsequent task performance? Truth re-enters through
consequence, not through correspondence. A false entry that consistently
produces failed actions loses its vitality through the utility signal. A
false entry that happens to produce successful actions in the short term
is harder to dislodge --- but that is not a design flaw, it is the
correct characterization of how human memory actually works. The
framework makes this dynamic visible and auditable rather than
pretending to escape it.

\begin{center}\rule{0.5\linewidth}{0.5pt}\end{center}

\hypertarget{the-accumulation-problem-reframed}{%
\subsection{3. The Accumulation Problem,
Reframed}\label{the-accumulation-problem-reframed}}

Storage cost, retrieval latency, and relevance noise are real
consequences of accumulation-only memory. Under some retrieval
architectures, larger corpora may degrade retrieval quality; under
better-indexed architectures, they may not. We do not treat this as the
primary motivation for the framework.

The primary motivation is that personal LLM memory systems are
personalizing retention in ways that can accumulate into drift, whether
or not their designers name the dynamic (descriptive claim, Section
1.1), and the design question is not ``how do we prevent drift'' but
``how do we mirror productively and compensate intelligently while the
drift happens.'' Naming the class is the precondition for giving it a
retention policy that serves its actual purpose rather than a policy
that pretends to serve a larger one. The operations that implement this
policy are specified in Section 5; before those operations can be
precisely defined, Section 4 names the objects they act on.

\begin{center}\rule{0.5\linewidth}{0.5pt}\end{center}

\hypertarget{system-model}{%
\subsection{4. System Model}\label{system-model}}

Before specifying operations, the framework names the objects those
operations act on and the states each object can occupy. This is the
system's object model --- analogous to RDF's abstract syntax or the data
model sections in Spanner and Dynamo. Every operation in Section 5 reads
from and writes to one or more of these entities. Naming them explicitly
is what allows conformance to be tested: if an implementation uses
different object boundaries or drops a required state transition, it is
not implementing this framework.

\hypertarget{core-entities}{%
\subsubsection{Core entities}\label{core-entities}}

\begin{longtable}[]{@{}
  >{\raggedright\arraybackslash}p{(\columnwidth - 6\tabcolsep) * \real{0.2500}}
  >{\raggedright\arraybackslash}p{(\columnwidth - 6\tabcolsep) * \real{0.2500}}
  >{\raggedright\arraybackslash}p{(\columnwidth - 6\tabcolsep) * \real{0.2500}}
  >{\raggedright\arraybackslash}p{(\columnwidth - 6\tabcolsep) * \real{0.2500}}@{}}
\toprule\noalign{}
\begin{minipage}[b]{\linewidth}\raggedright
Entity
\end{minipage} & \begin{minipage}[b]{\linewidth}\raggedright
Lifecycle states
\end{minipage} & \begin{minipage}[b]{\linewidth}\raggedright
Status flags
\end{minipage} & \begin{minipage}[b]{\linewidth}\raggedright
Required fields
\end{minipage} \\
\midrule\noalign{}
\endhead
\bottomrule\noalign{}
\endlastfoot
Raw buffer entry & pending → consolidated / rejected / expired & --- &
stable ID (content hash), ingestion timestamp, source pointer (Git blob
hash), origin channel, initial priority, candidate edge placeholders \\
Active wiki entry & active → decaying → archived & gravity-protected
(set by DECAY/AUDIT; orthogonal to lifecycle); quarantined (set by
CONSOLIDATE for low-cohesion entries; orthogonal to lifecycle) & ID,
commit hash, vitality score, gravity weight, quarantine flag,
last-accessed timestamp, cohesion bucket \\
Cold memory object & stored → recalled → re-compressed & --- & ID, Git
blob hash, original source URL or path, linkout commitment flag
(non-optional) \\
Audit record & created → closed & --- & entry ID, timestamp, suspension
result (degraded / unchanged / improved), outcome (restored /
gravity-reduced / archived) \\
Minority branch & open → promoted → closed & --- & Git branch reference,
incumbent entry ID, cluster size, contradiction edge count, cycles
open \\
\end{longtable}

\emph{Note: \texttt{gravity-protected} and \texttt{quarantined} are
status flags, not lifecycle stages. An entry can be \texttt{active} and
\texttt{gravity-protected} simultaneously. An entry can be
\texttt{decaying} and \texttt{quarantined} simultaneously. This matters
for conformance: conformance invariants that reference these flags apply
regardless of the entry's current lifecycle stage.}

\hypertarget{state-transition-rules}{%
\subsubsection{State transition rules}\label{state-transition-rules}}

Raw buffer entries are created by TRIAGE and consumed by CONSOLIDATE.
They do not transition backwards --- an entry either gets consolidated,
rejected on content grounds, or expires after the TTL window. Content is
immutable from the moment TRIAGE commits the entry. An entry in
\texttt{pending} state MUST be readable by CONSOLIDATE without
modification.

Active wiki entries are created by CONSOLIDATE and modified by DECAY and
AUDIT. They are never deleted. The \texttt{archived} state is terminal:
full content moves to cold memory, the index record remains with a
tombstone flag. This preserves audit history and prevents silent data
loss.

Cold memory objects are created by CONTEXTUALIZE when it produces a
depth-fitted working representation of an external source. The original
is immutable. Re-compression creates a new cold memory object and
updates the active wiki entry's commit hash --- the prior cold memory
object is retained, not overwritten.

Minority branches are created by CONSOLIDATE when a cluster of
mutually-supporting entries fails individual integration but warrants
preservation. Branches close only via two explicit paths: promotion,
when the cluster crosses the promotion threshold during a CONSOLIDATE
cycle; or AUDIT-triggered archival, when AUDIT confirms the incumbent
entry remains load-bearing and the branch has not grown across a defined
number of cycles. Branches are never closed silently --- if neither
condition has been explicitly evaluated and resolved, the branch remains
open.

\hypertarget{required-invariants}{%
\subsubsection{Required invariants}\label{required-invariants}}

\begin{itemize}
\tightlist
\item
  Every active wiki entry MUST maintain a valid commit hash pointing to
  its current content in Git
\item
  Every cold memory object MUST maintain a valid linkout to its original
  source --- this is the non-optional commitment CONTEXTUALIZE makes
  when it processes an external source during scheduled consolidation
\item
  TRIAGE MUST assign a content hash as the stable ID before writing to
  the buffer
\item
  Minority branches MUST NOT be closed silently --- closure requires
  either explicit promotion through CONSOLIDATE when the cluster crosses
  the promotion threshold, or explicit AUDIT-triggered archival when
  AUDIT confirms the incumbent remains load-bearing and the branch has
  not grown across a defined number of cycles
\item
  Audit records MUST be append-only --- no modification after creation
\item
  No operation MUST permanently delete any object --- terminal states
  are \texttt{archived} or \texttt{expired}, never hard-deleted
\end{itemize}

\begin{center}\rule{0.5\linewidth}{0.5pt}\end{center}

\hypertarget{companion-memory-the-framework}{%
\subsection{5. Companion Memory: The
Framework}\label{companion-memory-the-framework}}

\hypertarget{mapping-operations-to-mirror-and-compensate}{%
\subsubsection{5.0 Mapping operations to mirror and
compensate}\label{mapping-operations-to-mirror-and-compensate}}

Companion memory systems are analyzable across at least three layers.
The \emph{interaction/workflow layer} handles collection, annotation,
organization, and revisitation --- the territory of adaptive hypermedia
traditions that model and adapt to the individual user (Brusilovsky,
2001). The \emph{representation/retrieval layer} handles how memory
exists as text, summaries, embeddings, and links, and how it is
retrieved --- the territory of parametric and non-parametric memory
combinations in retrieval-augmented architectures {[}23{]}. The
\emph{retention/governance layer} determines what survives, how it is
protected, and when it is revised --- closer to lifecycle control and
architectural fitness than to retrieval alone {[}9{]}. This paper
operates primarily at the retention/governance layer, with CONTEXTUALIZE
reaching into the representation/retrieval layer to determine the form
in which external sources enter the wiki. The interaction/workflow layer
is out of scope: a companion memory system may have any workflow
interface, and that choice is independent of the retention policy this
paper theorizes.

Five operations and two supporting mechanisms implement the
retention/governance layer. Each is classified by which side of the
mirror-vs-compensate principle it implements. Throughout this section,
\emph{wiki} denotes the active knowledge store of any companion
knowledge system --- the term is borrowed from Karpathy's LLM Wiki but
is not limited to that implementation.

\begin{longtable}[]{@{}
  >{\raggedright\arraybackslash}p{(\columnwidth - 4\tabcolsep) * \real{0.3333}}
  >{\raggedright\arraybackslash}p{(\columnwidth - 4\tabcolsep) * \real{0.3333}}
  >{\raggedright\arraybackslash}p{(\columnwidth - 4\tabcolsep) * \real{0.3333}}@{}}
\toprule\noalign{}
\begin{minipage}[b]{\linewidth}\raggedright
Operation
\end{minipage} & \begin{minipage}[b]{\linewidth}\raggedright
Role
\end{minipage} & \begin{minipage}[b]{\linewidth}\raggedright
Mirror or compensate
\end{minipage} \\
\midrule\noalign{}
\endhead
\bottomrule\noalign{}
\endlastfoot
TRIAGE (streaming, ingestion) & Shallow filter; accept to buffer &
Mirror (neutral capture) \\
DECAY (continuous) & Survival-weighted retention on active wiki & Mirror
(operational continuity) \\
CONTEXTUALIZE (batched, scheduled) & Compress external sources to user's
working-context depth; preserve linkout & \textbf{Compensate} (selective
absorption) \\
CONSOLIDATE (batched, scheduled) & Deep coherence work, buffer-to-wiki
integration & \textbf{Compensate} (central mechanism) \\
AUDIT (slow cycle) & Structural stress test of high-gravity entries &
\textbf{Compensate} \\
Memory gravity & Load-bearing protection & Mirror \\
Minority-hypothesis retention & Variance preservation in buffer and
quarantine & \textbf{Compensate} \\
\end{longtable}

The compensate mechanisms are where the framework's safety story lives.
Their adequacy is discussed honestly in Sections 8.3 and 9.

\hypertarget{conflict-routing-matrix}{%
\paragraph{Conflict routing matrix}\label{conflict-routing-matrix}}

The mirror-vs-compensate principle is procedural rather than algorithmic
(§1.2): mirror by default in the streaming path, compensate during
scheduled integration windows, AUDIT as tiebreaker. At the case level,
this procedural rule has to decide how specific conflict types are
routed when mirror and compensate point in opposite directions. The
matrix below specifies routing for seven such cases. It states which
channel handles which conflict and under what rationale; it does not
specify friction coefficients, cycle counts for ``multi-cycle
accumulation,'' or quantitative thresholds for ``diverse sources.''
Those are calibration-dependent and belong in empirical follow-up work.

\textbf{Routing legend.} \emph{Mirror} = apply user-aligned behavior in
the interaction/working representation without mutating canonical
entries. \emph{Compensate} = route through CONSOLIDATE with elevated
friction relative to the default. \emph{Buffer} = store in minority
branch without immediate integration. \emph{AUDIT override} = route to
AUDIT with priority and apply the §5.8 gravity-reduction path if failure
persists. \emph{External correction} = flag for post-update CONSOLIDATE
review due to base-model prior change.

\begingroup\small
\setlength{\tabcolsep}{4pt}
\renewcommand{\arraystretch}{1.15}
\begin{longtable}{@{}>{\raggedright\arraybackslash}p{0.35cm}>{\raggedright\arraybackslash}p{2.8cm}>{\raggedright\arraybackslash}p{1.7cm}>{\raggedright\arraybackslash}p{4.8cm}>{\raggedright\arraybackslash}p{5.6cm}@{}}
\toprule
\textbf{\#} & \textbf{Conflict Type} & \textbf{Domain} & \textbf{Routing} & \textbf{Rationale} \\
\midrule
\endhead
1 & User vocabulary diverges from established ontology, no observed utility degradation & Operational & \textbf{Mirror in interaction and working representation; preserve divergence marker for later consolidation review.} Do not silently overwrite canonical wiki entries with the user's divergent vocabulary at streaming ingestion. & Continuity is the design goal in the operational domain, but ontology drift into the canonical store is an avoidable risk. The divergence marker gives CONSOLIDATE the context to decide whether to integrate the new vocabulary or preserve the established ontology at the next scheduled cycle. \\
\addlinespace
2 & User vocabulary diverges from established ontology, utility signal degrading across cycles & Operational $\rightarrow$ Epistemic & \textbf{Compensate.} Utility degradation triggers domain reclassification. CONSOLIDATE scores the divergence against the active wiki at higher friction. & The utility signal is the bridge: when a mirrored vocabulary produces worsening outcomes, the conflict has crossed into the epistemic domain. The routing change is not arbitrary --- it follows from the \S5.3 vitality dynamics making the operational justification for mirroring visibly weaker. \\
\addlinespace
3 & User repeatedly reinforces a consistent claim that external safety or epistemic signals (when available) flag as harmful, but user-reported utility remains stable or high & Safety & \textbf{Compensate regardless of utility signal.} Route to AUDIT priority queue; apply the highest available friction in the CONSOLIDATE path to any attempt to integrate reinforcing content; flag high-gravity reinforcing entries for stress test. Do not mirror at the operational layer even if continuity pressure is high. & This is the sycophancy failure mode. The utility signal is unreliable here because the user is reporting satisfaction with a harmful pattern the companion is reinforcing. Utility-based routing (the standard Operational $\rightarrow$ Epistemic bridge in row 2) does not trigger. Safety requires an explicit override: some reinforcement patterns must be resisted regardless of user-reported utility. The framework acknowledges this is a partial defense --- a fully novel harmful belief not flagged by any external signal falls into the residual failure mode named in \S9. \\
\addlinespace
4 & New evidence contradicts a high-gravity entry, single source, single cycle & Epistemic & \textbf{Default: Buffer.} Enter minority branch per \S5.7. Single-cycle single-source contradictions are treated as candidate signal rather than confirmed signal. \textbf{Exception:} implementations may define a high-trust source class whose single-cycle contradictions receive elevated initial weight; the framework does not specify the trust model, but the routing rule must not foreclose it. & The default posture protects against noise, but credible single-source discoveries --- a rigorous paper contradicting a belief, for example --- must not be structurally blocked by a universal single-source veto. The exception clause preserves implementation latitude without weakening the default. \\
\addlinespace
5 & New evidence contradicts a high-gravity entry, diverse sources, multi-cycle accumulation & Epistemic & \textbf{Compensate (consolidate candidate).} CONSOLIDATE evaluates promotion (\S5.5); if accumulated buffer pressure exceeds the incumbent's effective gravity under epistemic friction, integration occurs. & This is the canonical Prediction 4 case: multi-cycle diverse pressure is the mechanism by which minority positions become dominant. The routing cites \S5.5 explicitly so that the operational mechanism implementing this row is unambiguous. \\
\addlinespace
6 & High-gravity entry implicated in repeated poor outcomes over multiple AUDIT cycles & Epistemic + Safety & \textbf{AUDIT override.} Gravity is reduced through the \S5.8 gravity-reduction path regardless of centrality; the entry loses protection through the AUDIT channel rather than the CONSOLIDATE channel. & Continuity cannot indefinitely protect an entry that is producing bad outcomes. AUDIT is the structural defense against entrenchment of false load-bearing entries. This row is where the three-force architecture (\S5.6) earns its keep: gravity protects structure, utility signals through vitality, and AUDIT strips protection when structural prominence no longer tracks functional load. \\
\addlinespace
7 & Base model update introduces a factual prior contradicting a high-gravity wiki entry & Epistemic & \textbf{External correction channel.} The wiki entry is flagged for review on the next CONSOLIDATE cycle post-update. \textbf{This row depends structurally on architectural separability (\S8.3):} the external correction channel exists only because the wiki is not folded into base model weights. The separability commitment is what keeps this row operational across base model generations. & The base-model correction channel activates structurally; the wiki does not attempt to defend against the updated prior, but neither does it silently overwrite prior interpretations. Without separability, this row is impossible --- which is why separability is a safety commitment rather than an implementation detail. \\
\bottomrule
\end{longtable}
\endgroup

\textbf{Limitation.} Row 7 names the base-model correction channel, but
the structural residual --- fully novel bad beliefs not represented in
the base model and not contradicted by subsequent experience --- is not
captured by any row in this matrix. This is the limit acknowledged in §9
and not resolved by the routing logic. The matrix specifies how the
framework behaves when the relevant conflict signal exists; it does not
manufacture signal that external sources do not provide. The matrix does
not define calibration parameters (e.g., cycle counts, source diversity
thresholds), which are explicitly left to implementation and empirical
validation.

\hypertarget{raw-buffer-and-consolidation-cycle}{%
\subsubsection{5.1 Raw buffer and consolidation
cycle}\label{raw-buffer-and-consolidation-cycle}}

The framework splits ingestion from integration. A raw buffer accepts
entries as they arrive; the active wiki is modified only during
scheduled consolidation cycles.

This is the sleep-function architecture. Streaming ingestion writes to
the buffer through a shallow TRIAGE filter. Deep coherence work ---
classification, contradiction resolution, integration with the active
wiki, promotion of minority positions, flagging of gravity conflicts ---
runs during a batched CONSOLIDATE operation on a scheduled rhythm
(nightly, weekly, or event-driven depending on the use case). The buffer
is short-term; the active wiki is long-term; consolidation is the
bridge.

The reason to split is that streaming coherence is self-sealing. A
single entry arriving alone and scored against the dominant wiki gets
quarantined immediately if it contradicts the dominant interpretation,
which means the dominant interpretation never updates. Batched
consolidation breaks this lock. Multiple buffer entries are scored
against each other as well as against the wiki, and three
mutually-supporting entries against a high-gravity wiki entry is a
different signal from one isolated contradiction. The minority position
can accumulate buffer pressure and shift the dominant interpretation
during a consolidation window. This is how bounded agents actually
update when they update at all.

\hypertarget{triage-streaming-shallow-filter}{%
\subsubsection{5.2 TRIAGE --- streaming shallow
filter}\label{triage-streaming-shallow-filter}}

TRIAGE runs at ingestion. Its job is deliberately limited: reject
obvious garbage, deduplicate against the recent buffer, check structural
validity, assign an ingestion timestamp. That is all. TRIAGE does not
classify entries into cohesion buckets, does not score them against the
active wiki, and does not make retention decisions. Everything that
passes TRIAGE enters the raw buffer and waits for the next consolidation
cycle.

Keeping TRIAGE shallow is a design commitment. The moment TRIAGE starts
doing coherence work, the architecture collapses back to streaming and
the self-sealing problem returns.

\hypertarget{decay-survival-weighted-retention-on-the-active-wiki}{%
\subsubsection{5.3 DECAY --- survival-weighted retention on the active
wiki}\label{decay-survival-weighted-retention-on-the-active-wiki}}

DECAY runs continuously over the active wiki. Every entry carries a
vitality score reflecting its contribution to the user's useful outcomes
plus its structural role in the wiki's own coherence:

\begin{verbatim}
vitality(entry) =
    recency_weight   * (1 / days_since_access)
  + frequency_weight * access_count
  + utility_weight   * task_predictive_utility(entry)
  + gravity_weight   * memory_gravity(entry)
  - wear_penalty     * summarization_distortion(entry)
\end{verbatim}

\texttt{task\_predictive\_utility} is the utility signal: did acting on
this entry produce results the user judged useful? Under the
mirror-vs-compensate principle, this is a mirror mechanism. The system
is supposed to retain what serves the user operationally. What prevents
the vitality function from collapsing into pure satisfaction-chasing is
the gravity term --- entries load-bearing for the wiki's own coherence
are protected even when rarely useful in the direct sense.

Entries below the vitality threshold are compressed into summary form
rather than deleted. Information degrades gracefully. Vitality-driven
compression does not override the gravity-protection floor (§7.5): an
entry whose G\_i\^{}base remains above the floor is protected from
compression even if its vitality score falls below threshold, on the
same structural-centrality grounds that protect it from DECAY. This
preserves the quiet-foundations guarantee across both retention
mechanisms.

\hypertarget{contextualize-depth-fitted-compression-of-external-sources}{%
\subsubsection{5.4 CONTEXTUALIZE --- depth-fitted compression of
external
sources}\label{contextualize-depth-fitted-compression-of-external-sources}}

CONTEXTUALIZE is the framework's response to a problem the other
operations do not handle: external sources do not have a single
canonical compression. The same architecture decision record yields a
different useful summary for a Product Owner than for a Developer
reading it the same afternoon. The Product Owner needs goals, tradeoffs,
and stakeholder rationale; the Developer needs implementation
constraints, library choices, and edge cases. Neither is wrong. Both are
\emph{contextually correct compressions} of the same artifact, fitted to
the working depth at which the user is currently engaging the topic.

A naive ingestion pipeline that compresses external sources to some
imagined complete representation pays a double cost. It bloats the wiki
with content the user does not need, and it makes the wiki harder to
consolidate against because the entries are too long to participate
cleanly in coherence operations. A pipeline that compresses too
aggressively pays a different cost: it strips out the depth the user
actually needs and leaves them with summaries that are accurate but
operationally useless. CONTEXTUALIZE proposes a third path. Compress
external sources to fit the user's current working-context depth on the
relevant topic, \emph{and preserve a linkout to the full external
source} so the original is recoverable when context shifts.

The closest prior art is D-Mem (You, Yuan \& Cai, arXiv 2603.18631),
which proposes a dual-process memory architecture: a lightweight
retrieval path for fast inference and a Full Deliberation module that
falls back to processing raw dialogue history when retrieval quality
falls below a learned Quality Gating threshold. D-Mem thus keeps
fast-path compression alongside access to the full original content,
with the gating policy deciding which path to use at retrieval time.
CONTEXTUALIZE differs in two ways: it decides the appropriate
compression depth \emph{before} integration rather than switching
reactively at retrieval time, and it infers that depth from the user's
wiki topology and query behavior rather than from the retrieval query
alone. The coordination bundle --- depth inference from user context,
compression fitted to that depth, originals preserved as a structural
non-optional commitment, deferred to the dream cycle rather than run at
streaming ingestion --- is what prior work does not assemble together.
D-Mem provides one of the components; CONTEXTUALIZE provides the
governance logic that determines when and how deeply to compress.

Two design commitments matter here. First, CONTEXTUALIZE runs in the
dream cycle, not at runtime ingestion. Streaming ingestion (TRIAGE) is
intentionally shallow because runtime cost has to stay low;
depth-fitting compression is expensive enough that batching it with
sleep consolidation is the right place architecturally. The raw external
source survives in the buffer until the next dream cycle, which is also
a safety property: if the user's context shifts between ingestion and
consolidation --- they leave the Product Owner role and become an
Engineering Manager, or they shift from feature work to architecture ---
the next dream cycle compresses against the \emph{new} context rather
than the old one. The buffer's job between cycles includes preserving
the option to re-compress.

Second, the depth is \emph{inferred} by default, not explicitly set by
the user. Asking the user to specify their working-context depth on
every ingested source is operationally absurd; the cognitive load
defeats the purpose of the system. The companion infers depth from the
user's other wiki entries, recent query patterns, and the topical
neighborhood the source falls into. This is the same inference move the
framework already relies on for memory gravity (which entries are
load-bearing for the user's current wiki structure) and for
minority-hypothesis retention (which contradictions might matter to this
user later). Inferred depth fails sometimes, in ways the user cannot
easily catch --- which is why the linkout to the full source is
non-optional. The user can always go back to the original. The
compressed entry is the working representation, not the truth.

The metabolic analogy is selective absorption. Biological cells do not
absorb everything in their environment; they absorb what their current
metabolic state can use, and the rest passes through or is excreted. The
companion memory equivalent is that wiki ingestion should compress
external sources to the user's current depth-of-engagement with the
topic, not to some imagined complete or context-free representation. The
wiki is metabolically active; what counts as nutritious input depends on
what the user is currently digesting.

This introduces a third storage tier beyond the raw buffer and the
active wiki: \emph{cold memory}. The raw external source --- the full
document, article, or record that TRIAGE accepted into the buffer ---
does not need to live in the active wiki after CONTEXTUALIZE has
produced the depth-fitted working representation. But it should not be
deleted either, because CONTEXTUALIZE's mandatory linkout commitment
means the user must be able to retrieve it when context shifts. Cold
memory is the named destination for these originals: high-capacity,
low-access-frequency storage that holds the sources the wiki has already
processed. Entries in cold memory are not retrieved in normal operation;
they are retrieved only when a user or CONSOLIDATE cycle determines that
the working representation needs to be re-compressed against a new
context. The three-tier architecture --- cold memory (originals), raw
buffer (pending consolidation), active wiki (depth-fitted working
representations) --- makes the storage obligations explicit rather than
leaving them as an implicit consequence of linkout preservation.

We treat CONTEXTUALIZE as a fifth operation rather than a sub-step of
TRIAGE because the work it does is substantively different: TRIAGE is a
gatekeeper for the raw buffer, CONTEXTUALIZE is a depth-fitting
compressor for external sources during scheduled integration. They run
on different timescales (streaming vs batched), they touch different
content types (everything vs external sources specifically), and they
have different failure modes (TRIAGE fails by letting noise in;
CONTEXTUALIZE fails by compressing at the wrong depth). The
architectural separation matters for the same reason TRIAGE and
CONSOLIDATE were separated in the first place: collapsing them
re-introduces the streaming-coherence trap.

\hypertarget{consolidate-batched-deep-integration}{%
\subsubsection{5.5 CONSOLIDATE --- batched deep
integration}\label{consolidate-batched-deep-integration}}

CONSOLIDATE is the framework's central compensate mechanism. It runs on
a schedule against the raw buffer and the active wiki together.

The operation has four phases:

\begin{enumerate}
\def\labelenumi{\arabic{enumi}.}
\tightlist
\item
  \textbf{Buffer-internal scoring.} Each buffer entry is scored against
  other buffer entries for mutual support and contradiction,
  independently of the active wiki. This is where accumulated minority
  pressure becomes visible.
\item
  \textbf{Wiki scoring.} Each buffer entry is also scored against the
  active wiki for semantic consistency, task alignment, and
  contradiction cost with high-gravity nodes.
\item
  \textbf{Classification and routing.} Entries are placed on a fuzzy
  coherence gradient using trapezoid membership functions {[}49{]}.
  High-cohesion entries integrate directly. Mid-cohesion entries flag
  for future attention. Low-cohesion entries quarantine for
  re-evaluation in future cycles.
\item
  \textbf{Minority-pressure promotion.} Entries that contradict the
  active wiki \emph{individually} but mutually support each other
  \emph{in the buffer} are flagged as candidate updates to the dominant
  interpretation rather than quarantined. The consolidation cycle either
  updates the wiki to accommodate them, or leaves them in extended
  quarantine for the next cycle to re-evaluate with additional buffer
  context. Single contradictions are treated as noise; accumulated
  contradictions are treated as signal.
\end{enumerate}

This is the structural answer to the critique that stored variance is
not effective variance. Under streaming coherence filtering, a minority
hypothesis is stored but never integrated. Under batched CONSOLIDATE
with minority-pressure promotion, a minority hypothesis has a real
mechanism for becoming the dominant one when enough supporting evidence
accumulates within a single consolidation window.

Batched consolidation can also amplify correlated noise rather than
genuine correction. Multiple mutually-supporting but mistaken entries
may accumulate enough buffer pressure to challenge a correct dominant
interpretation, precisely because the mechanism rewards mutual support.
CONSOLIDATE is therefore anti-entrenchment, not truth-guaranteeing.
Whether buffer-internal agreement tracks genuine correction more often
than clustered error is an empirical question the framework does not
answer, and the gap is explicitly open follow-up work. Source diversity
weighting, time-spread requirements between supporting entries, and
external validation signals are candidate defenses we flag without
currently implementing.

The framework also acknowledges a \emph{Valley of Amnesia} during
paradigm shifts: if the promotion threshold is crossed prematurely, the
system risks a sudden loss of operational continuity before the new
interpretation has achieved structural stability. This is a
transition-stability concern distinct from the correlated-noise concern
above, though the two share a mechanism and are both open empirical
questions the framework does not resolve.

\hypertarget{memory-gravity-load-bearing-for-operational-continuity}{%
\subsubsection{5.6 Memory gravity --- load-bearing for operational
continuity}\label{memory-gravity-load-bearing-for-operational-continuity}}

Memory gravity protects entries whose removal would cascade through the
knowledge base. Not because those entries are true, but because they are
structurally essential for the wiki's coherent operation.

Define the active wiki as a weighted directed graph W = (V, E), where
vertices are wiki entries and edges are dependency references between
them. For entry i ∈ V, base gravity is defined as a function:

\begin{verbatim}
G_i^base = f(C(i), F(i))
\end{verbatim}

where:

\begin{itemize}
\tightlist
\item
  \textbf{C(i)} is a centrality measure over W capturing how much of the
  wiki's reference structure flows through i, directly and transitively.
  The framework is agnostic about the specific implementation:
  eigenvector centrality, PageRank, or a domain-appropriate variant are
  all valid choices, so long as the chosen measure is documented.
\item
  \textbf{F(i)} is downstream fragmentation cost: a prospective measure
  of how much of the wiki's coherent operation would break if i were
  removed at the current moment.
\end{itemize}

G\^{}base MUST satisfy four properties:

\begin{enumerate}
\def\labelenumi{\arabic{enumi}.}
\tightlist
\item
  \textbf{Monotonicity in centrality.} If C(i) \textgreater{} C(j) and
  F(i) = F(j), then G\_i\^{}base \textgreater{} G\_j\^{}base.
  Structurally more central entries receive greater protection, all else
  equal.
\item
  \textbf{Monotonicity in fragmentation.} If F(i) \textgreater{} F(j)
  and C(i) = C(j), then G\_i\^{}base \textgreater{} G\_j\^{}base.
  Entries whose removal would cause greater damage receive greater
  protection, all else equal.
\item
  \textbf{Sub-linear growth under incumbency.} As C(i) grows without
  bound --- where centrality is instantiated via in-degree or any
  monotone incumbency proxy --- G\^{}base MUST grow sub-linearly in that
  quantity. This prevents the \emph{Absolute Incumbency Trap}: a failure
  mode in which a maximally-referenced entry accumulates unbounded
  protection and becomes structurally unassailable regardless of how
  much contradicting evidence subsequently accumulates against it.
  Sub-linear growth is a safety property, not an optimization. It
  complements property 4 by constraining the \emph{shape} of gravity's
  response to increasing incumbency, not merely the range of gravity
  values.
\item
  \textbf{Boundedness.} G\^{}base is normalized to a bounded range. The
  specific range is implementation-defined; the required property is
  that gravity values be normalized to a bounded scale and remain
  comparable across entries within the same wiki state. Boundedness
  alone would permit linear accumulation toward the ceiling followed by
  a cliff, which is why property 3 is independently required.
\end{enumerate}

\textbf{Time-decay component.} Effective gravity incorporates a decay
factor reflecting elapsed time since last access:

\begin{verbatim}
G_i^eff(t) = G_i^base · D(t − t_last_access)
\end{verbatim}

where G\_i\^{}base is evaluated against the current wiki topology at
time t (C and F are computed against the wiki as it stands now), and
D(Δt) is a decay function with three required properties: D(0) = 1, D is
monotonically non-increasing in Δt, and D approaches zero as Δt grows
without bound. The specific functional form (exponential, power-law,
piecewise) is implementation-dependent and must be documented. The
framework does not prescribe one.

The decay function modulates effective gravity but does not govern
decay-eligibility. The gravity-protection floor (§7.5) is evaluated
against G\_i\^{}base --- an entry whose structural centrality remains
high stays above the decay-eligibility boundary regardless of how long
since last access. Time-decay reduces G\_i\^{}eff (affecting vitality
through §5.3) but cannot drive a structurally central entry below the
§7.5 floor. This is how quiet foundations survive: gravity's structural
component does not decay, only the access-modulated effective component
does.

\textbf{Architectural note on separation of forces.} Base gravity
G\_i\^{}base is a \emph{pure structural property} of the wiki graph;
effective gravity G\_i\^{}eff is its access-modulated form. The utility
signal \texttt{task\_predictive\_utility} (§5.3) does not enter either.
This preserves the framework's three-force architecture: gravity
protects structurally load-bearing entries, utility drives
vitality-based decay through §5.3, and AUDIT (§5.8) is the mechanism
that strips protection from high-gravity entries revealed as
non-load-bearing through counterfactual suspension. Folding utility into
effective gravity would collapse two distinct mechanisms into one and
would change the framework's compensate story; the three forces remain
distinct. When gravity appears as a term in the vitality formula (§5.3),
the value referenced is G\_i\^{}eff --- vitality is evaluated at time t
and inherits gravity's time-decay.

\textbf{A known failure mode.} A false entry that became load-bearing
before it was recognized as false is \emph{more} protected, not less.
The framework does not eliminate this. AUDIT is the defense: a
gravity-protected entry that fails consequence tracking under repeated
AUDIT cycles loses its protection through the §5.8 gravity-reduction
path, not through the gravity formula itself. This works only if AUDIT
is sensitive enough to detect the failure, which is an open problem we
name in §9. We do not claim the framework prevents entrenchment of false
beliefs. We claim it localizes entrenchment to a specific failure mode
and gives that failure mode a named defense that can be improved
independently.

Memory gravity borrows from architectural gravity in software systems
{[}9{]}: certain decisions become load-bearing because too much is built
on top of them, and removing them is costly regardless of whether they
were ideal choices.

The gravity mechanism has prior art in bibliometrics and citation
analysis worth acknowledging directly. PageRank {[}32{]} computes node
importance in a directed graph by iteratively propagating authority from
high-degree nodes to their neighbors, which is one valid implementation
of C(i) above. The h-index {[}12{]} measures impact by the largest h
such that h papers have each been cited at least h times: a
threshold-based gravity model that protects a core of highly-cited work
while allowing lower-impact work to fade. Both operate on retrospective
citation graphs: importance is measured by what has historically
referenced the node. Memory gravity differs on a prospective dimension
that bibliometrics does not address: F(i) measures what would break if
the entry were removed now, not what has historically referenced it.
This is gravity by structural consequence rather than historical
popularity. The distinction matters because a recently integrated entry
may still be structurally critical to the user's current working context
even if few entries reference it yet; a purely retrospective measure
would underprotect it.

\hypertarget{minority-hypothesis-retention-variance-preservation}{%
\subsubsection{5.7 Minority-hypothesis retention --- variance
preservation}\label{minority-hypothesis-retention-variance-preservation}}

Dormant alternatives are kept in the buffer and in quarantine at low
storage cost. They are not stored for their own sake; they are stored so
that the next consolidation cycle has something to score against
incoming entries. Without minority-hypothesis retention, the buffer
arrives at each consolidation cycle with no historical variance, and
minority positions must accumulate their full support from scratch
within a single cycle window. With it, a minority position that started
building buffer pressure three cycles ago can complete the accumulation
in the current cycle.

Controlled variance injection is a standard pattern in recommender
systems and reinforcement learning. We apply it to companion memory
without claiming the memory case reduces to the recommender case --- the
goals are different (epistemic variance versus engagement variance), but
the mechanical move is the same.

A natural extension for future work: the quarantine structure for
minority hypotheses could be formalized as named branches in a
version-controlled knowledge store, with merge semantics governing
integration --- giving users an inspectable mental model for
minority-hypothesis state beyond what the word ``quarantine'' alone
conveys.

\hypertarget{audit-structural-stress-test}{%
\subsubsection{5.8 AUDIT --- structural stress
test}\label{audit-structural-stress-test}}

AUDIT runs on a slow cycle (monthly or longer). It temporarily suspends
the highest-gravity entries and observes whether query performance
degrades:

\begin{verbatim}
FOR each entry in top_N_by_gravity:
    suspend from active wiki
    run N queries that previously accessed this entry
    IF query performance degrades:   restore, confirm gravity
    IF query performance unchanged:  reduce gravity — entry is dead weight
    IF query performance improves:   archive — entry was actively interfering
\end{verbatim}

AUDIT is not a truth-correction mechanism. It tests whether high-gravity
entries are still load-bearing for the agent's current operation. An
entry that was once central but that the agent has outgrown becomes dead
weight. An entry that was actively interfering with current queries gets
archived.

The framework invites a Kuhnian reading worth making explicit (Kuhn,
1962). Kuhn's account of normal science describes exactly the failure
mode that AUDIT is designed to interrupt: a paradigm becomes
self-reinforcing precisely because it organizes how new evidence is
interpreted, and anomalies that would otherwise force a paradigm shift
are routed to the periphery as exceptions, measurement errors, or open
problems ``for future work.'' A wiki without AUDIT is structurally
analogous to normal science without crisis pressure --- accumulating
coherence at the cost of accumulating debt against unaddressed
anomalies. AUDIT does not resolve the Kuhnian tension; it makes the cost
of unaddressed anomalies visible at the entry level rather than letting
it accumulate invisibly at the wiki level.

A recent empirical operationalization of Kuhnian paradigm structure in
Wikipedia concept networks {[}18{]} treats paradigms as structural
modules with temporal signatures, detecting periods of structural
stability and instability through the graph topology of article
hyperlinks. AUDIT extends this logic to the personal companion memory
context with a periodic, \emph{performance-based} stress-test: rather
than identifying paradigm structure retrospectively through historical
analysis, it suspends candidate high-gravity entries and measures
whether their removal degrades current agent performance. The result is
a Kuhnian discipline that operates at the entry level rather than across
decades --- and that makes the cost of accumulated anomaly debt visible
\emph{before} a forced paradigm shift, not after.

AUDIT sensitivity is an open problem. If the query set used for stress
testing is narrow or self-confirming, harmful central nodes stay
protected. We do not solve this. We flag it as the specific direction
where the compensate side of the framework most needs work.

A suggestive analogue from a structurally adjacent domain: Men et al.'s
ShortGPT (arXiv 2403.03853) shows that transformer layers with low Block
Influence scores --- measuring how much a layer transforms its inputs
relative to its outputs --- can be removed with minimal performance
loss, even when those layers are structurally central. The result
inverts naive gravity: positional centrality in a sequential network
does not reliably predict functional necessity. AUDIT makes the same
move at the knowledge-entry level. A high gravity score is a structural
proxy for functional importance, not a measurement of it. Only empirical
suspension reveals the difference. The analogy is not mechanistically
exact --- transformer layer redundancy differs from semantic entry
centrality --- but it provides independent evidence that the gap between
structural prominence and functional load is a real phenomenon in at
least one adjacent system class.

\begin{center}\rule{0.5\linewidth}{0.5pt}\end{center}

\hypertarget{integration-with-existing-systems}{%
\subsection{6. Integration with Existing
Systems}\label{integration-with-existing-systems}}

The framework is additive. Karpathy's LLM Wiki provides the incremental
compilation pattern; the companion layer adds a raw buffer tier, a
consolidation schedule, and metadata files for gravity and cohesion
tracking. MemPalace provides the retrieval architecture; the companion
layer operates at the entry level underneath. LLM Wiki v2's Ebbinghaus
decay is a special case of the vitality function with utility and
gravity weights set to zero.

The companion framing does not change these integration patterns. It
changes what the combined system is \emph{for}: a retention policy
serving a specific user under the mirror-vs-compensate principle, not a
general-purpose memory architecture.

A note on storage representation. The framework is agnostic about how
wiki content is encoded under the hood --- plain markdown, dense
embeddings, or a hybrid of both. In practice, we recommend the hybrid
default that existing RAG systems already use: plain text as the
authoritative content store plus embeddings as a retrieval index
alongside it. This recommendation rests on a design intuition rather
than a fully argued proof. The external correction channel described in
Section 8.3 --- base-model updates providing free reinterpretation
without the framework implementing them --- depends on wiki content
surviving a base-model swap without being silently rewritten. Plain text
as primary representation makes that survival auditable in a way
embeddings-only storage does not, because a model swap forces
re-embedding during which meaning can shift in ways the user cannot
inspect after the fact. Because model swaps also invalidate existing
embedding indices --- different base models have different embedding
spaces and semantic geometries --- the framework assumes a re-indexing
pass over the plain-text wiki content is a required step of any
base-model generation update; the embedding layer is a derived artifact,
while the plain-text layer is the preserved one. We treat this as a
design commitment worth flagging, not as a proof that hybrid systems
cannot preserve stable semantics across model swaps. The empirical
question of whether plain text is necessary or merely convenient for
cross-generation auditability is left open for future work.

\hypertarget{implementation-commitments}{%
\subsubsection{6.1 Implementation
commitments}\label{implementation-commitments}}

The framework makes seven architectural commitments beyond the storage
representation note above. The first two --- two-scheduler architecture
and homeostasis-driven sleep scheduling --- are conformance-level: an
implementation that violates them breaks the governance contract. The
remaining five are principled reference-implementation recommendations
grounded in the governance obligations; alternative implementations
satisfying the conformance invariants in Section 7.5 remain valid even
if they differ here.

\textbf{Two-scheduler architecture.} The runtime splits into two
schedulers with distinct latency requirements. The hot-path scheduler
gates TRIAGE and retrieval reads on the conversation thread --- these
must complete within the user's conversational latency budget. The
sleep-cycle scheduler queues DECAY, CONSOLIDATE, and AUDIT as background
jobs. Conflating these into a single scheduler is the most common naive
implementation failure: sleep-cycle operations steal latency from the
hot path, or hot-path latency requirements cause sleep-cycle operations
to be perpetually deferred. The two schedulers are a conformance
boundary: an implementation that blocks conversation during CONSOLIDATE
is non-conforming regardless of correctness.

\textbf{Sleep scheduling by homeostasis layer, not cron.} Sleep-cycle
operations are scheduled by a homeostasis layer reading a state vector
--- interaction recency, thermal state, battery trajectory, storage
pressure, and available signals about the user's next active session
window. If the system can infer that a user leaves at 8 AM and
consolidation requires 30 minutes, 4:45 AM is the computed window, not a
fixed nightly timer. This is the difference between reactive scheduling
(run at a fixed time) and anticipatory homeostasis (run when conditions
are optimal given the system's forward projection). The closest
published template is the device-state-constrained scheduling in
Bonawitz et al.'s federated learning system design (arXiv:1902.01046),
which constrains background work to idle/charging/unmetered conditions
with explicit abort semantics when conditions change.

\textbf{Persistence: content-addressable store plus relational metadata
index.} The persistence layer splits into two components with distinct
roles. A content-addressable filesystem --- canonical implementation:
Git --- stores full wiki entry content and version history. A relational
metadata index --- canonical implementation: PostgreSQL --- stores
operational state only: entry tokens, vitality scores, gravity weights,
contradiction and dependency edges, audit records. Three tables:
\texttt{entries} (entry ID, commit hash, vitality, gravity, quarantine
flag, last-accessed), \texttt{edges} (source, target, edge type,
weight), \texttt{audit\_log} (entry ID, timestamp, result, outcome). The
primary index for DECAY performance is vitality score --- DECAY is the
highest-frequency operation and a full table scan without this index is
the first performance bottleneck in any realistic implementation.

\textbf{Four communication boundaries.} The architecture has four
distinct communication surfaces, each with a different right answer:

\begin{longtable}[]{@{}
  >{\raggedright\arraybackslash}p{(\columnwidth - 4\tabcolsep) * \real{0.3333}}
  >{\raggedright\arraybackslash}p{(\columnwidth - 4\tabcolsep) * \real{0.3333}}
  >{\raggedright\arraybackslash}p{(\columnwidth - 4\tabcolsep) * \real{0.3333}}@{}}
\toprule\noalign{}
\begin{minipage}[b]{\linewidth}\raggedright
Boundary
\end{minipage} & \begin{minipage}[b]{\linewidth}\raggedright
Approach
\end{minipage} & \begin{minipage}[b]{\linewidth}\raggedright
Rationale
\end{minipage} \\
\midrule\noalign{}
\endhead
\bottomrule\noalign{}
\endlastfoot
Internal runtime components & Direct function calls / Unix sockets & No
protocol overhead; fixed known interface \\
Storage layer & Filesystem ops + SQL queries & Standard, debuggable, no
network hop \\
Federation (device to device) & gRPC with proto-enforced schema & Proto
schema enforces anonymized-outbound-only at protocol level \\
External model APIs & Provider-dependent HTTPS REST & Provider-dictated;
not our choice \\
\end{longtable}

The gRPC choice at the federation boundary warrants explicit
justification. Most federated system privacy enforcement relies on
platform isolation, cryptographic aggregation, or protocol-level
restrictions rather than schema-level constraints. One possible approach
--- proposed here as a design commitment rather than a novelty claim ---
is to use the proto schema itself as the privacy boundary: if personally
identifying content cannot be expressed in the anonymized-outbound-only
schema, it cannot cross the device boundary regardless of what the
software layer attempts. This makes the schema the enforcement surface,
not a constraint layered on top of enforcement. Whether prior systems
have used this exact mechanism is not established by a systematic search
and should not be treated as a settled absence-of-prior-art claim.

\textbf{No Kafka for the single-user base case.} Kafka solves
high-throughput multi-producer event streaming at distributed scale. A
single user's conversation stream does not generate that volume, and
Kafka introduces latency, operational overhead, and infrastructure
complexity that are unjustified here. Kafka becomes defensible at the
federation ingestion bus --- when multiple companion instances sync
anonymized gravity and minority signals --- as one valid choice among
several durable streaming substrates. It belongs in the federation spec,
not the base spec.

\textbf{CLI over MCP for internal tooling.} In a local-first single-user
system with a fixed known tool set, MCP's dynamic tool discovery adds
token overhead without benefit. The reasoning layer has a fixed set of
operations: search, commit, fetch, audit. Direct CLI invocation is
leaner, more debuggable, and does not spend tokens on schema negotiation
and tool listing on each session. MCP becomes relevant when the tool
ecosystem is unknown at design time or shared across third parties ---
neither applies here.

\textbf{Hardware as constraint surface, not ceiling.} The spec states
constraint priorities --- local-first, data boundary, user sovereignty
--- without committing to a capability ceiling for the reasoning layer.
The reasoning layer's capability is a function of the hardware
generation available at deployment time. CONSOLIDATE and AUDIT require
LLM-grade semantic reasoning; what that means in practice scales with
what runs on-device at deployment time. A spec that committed to a
specific parameter count in 2026 would be outdated within 18 months. The
governance contract is hardware-generation independent: sovereignty
obligations are invariant, capability adapts to the hardware available
at deployment.

\begin{center}\rule{0.5\linewidth}{0.5pt}\end{center}

\hypertarget{predictions-and-proxies}{%
\subsection{7. Predictions and Proxies}\label{predictions-and-proxies}}

Every claim below is a prediction with a proxy measure. None is a claim
about retrieval quality or truth tracking.

\textbf{Prediction 1 --- Coherence stability under drift.} A companion
system with gravity protection and batched consolidation maintains
internal cross-reference integrity and stable self-description entries
under simulated user drift over multiple months. \emph{Proxy:}
cross-reference integrity score plus stability of designated
self-description entries across simulated sessions. \emph{Test:}
simulate a user whose preferences shift gradually over N months; compare
gravity-protected consolidation against vitality-only pruning.

\textbf{Prediction 2 --- Fragility resistance.} Gravity-protected
vitality pruning produces fewer orphaned entries and broken references
under storage pressure than vitality-only pruning. \emph{Proxy:} count
of orphaned entries and broken cross-references after successive pruning
cycles on a knowledge base with known dependency structure. \emph{Test:}
apply both pruning strategies to identical seeded corpora under
equivalent storage caps.

\textbf{Prediction 3 --- Monoculture resistance.} Active consolidation
with minority-pressure promotion maintains higher entropy in access
distribution and higher conceptual diversity in surfaced entries than
pure vitality-ranked retrieval. \emph{Proxy:} Shannon entropy of access
distribution over N weeks, plus semantic diversity of entries surfaced
in response to a fixed probe query set. \emph{Test:} run paired systems
with and without minority-pressure promotion; measure divergence in
entropy and diversity over time.

\textbf{Prediction 4 --- Effective minority-hypothesis influence.} This
is the sharpest prediction and the one the framework is most willing to
be falsified on. The claim is not that minority hypotheses are
\emph{stored}, nor that they are \emph{surfaced}, but that they
measurably \emph{change downstream outputs} at a non-trivial rate.
\emph{Proxy:} the resurfacing-to-influence rate --- for each minority
entry promoted during a consolidation cycle, measure whether the
system's responses to subsequent relevant queries differ from responses
the same system would have produced without the promotion. \emph{Test:}
run the system in paired modes (consolidation with minority-pressure
promotion active vs.~inactive) against an identical query stream;
measure output divergence conditional on minority-entry involvement.

A critical distinction from existing benchmarks: LongMemEval's
``knowledge updates'' category (Shi et al., arXiv 2410.10813) evaluates
whether a system correctly incorporates \emph{explicit} profile changes
--- a user states their city has changed, and the system should
remember. Prediction 4 targets a structurally different failure mode. A
minority hypothesis, as defined in this framework, is a cluster of
mutually-supporting entries that (a) contradict a high-gravity incumbent
entry individually, such that no single entry achieves integration on
its own; (b) cannot achieve integration in any single consolidation
cycle alone because their combined pressure falls below the promotion
threshold; but (c) can achieve integration through multi-cycle buffer
pressure accumulation, when the cluster builds across successive
consolidation windows against a centrality-protected incumbent. This is
not a knowledge update. It is a belief revision that happens
structurally through the consolidation architecture rather than through
explicit user input or a direct instruction to update. No existing
benchmark, including LongMemEval, tests this structural failure mode ---
which is why the framework is prepared to be falsified on it
independently of retrieval accuracy results.

A second distinction is from TeaFarm, the counterfactual evaluation
scheme embedded in THEANINE (Ong et al., arXiv:2406.10996). TeaFarm
tests whether memory changes outputs by comparing system responses with
and without specific memory entries --- a counterfactual
presence/absence design. It is the strongest existing evaluation of
memory influence and it must be engaged directly. Prediction 4 targets a
structurally different failure mode. TeaFarm asks: \emph{does memory
change output?} Prediction 4 asks: \emph{does a cluster of
individually-quarantined contradictions achieve integration through
multi-cycle buffer pressure accumulation against a centrality-protected
incumbent, and does that integration measurably shift downstream
outputs?} A system can pass TeaFarm's counterfactual test --- memory
demonstrably changes output --- while still exhibiting the failure mode
Prediction 4 targets: the specific cluster that should have triggered
belief revision was quarantined entry-by-entry across successive
consolidation windows and never integrated, because no single entry
crossed the promotion threshold alone. The failure mode is
architectural, not retrieval-based. TeaFarm does not test for it. The
prediction is falsifiable independently of TeaFarm results, and the test
protocol differs: instead of presence/absence comparison, Prediction 4
requires tracking multi-cycle buffer accumulation, measuring cluster
size against centrality thresholds, and comparing paired consolidation
runs with and without minority-pressure promotion active.

Retention alone is not variance. Surfacing alone is not variance. Only
\emph{influence on a downstream output} is effective variance, and that
is what Prediction 4 is willing to be measured on.

\hypertarget{conformance}{%
\subsubsection{7.5 Conformance}\label{conformance}}

This section states normative invariants for each operation. An
implementation that violates any of these is not conforming to this
framework, regardless of how it names its components. The invariants are
the spec's enforcement surface --- the point at which ``here is a
persuasive framework'' becomes ``here is something a builder can test
against.''

\textbf{TRIAGE} - MUST NOT perform semantic contradiction resolution ---
that work belongs to CONSOLIDATE - MUST assign a stable content-hash ID
and ingestion timestamp before writing to the buffer - MUST be
idempotent on identical input content: same content hash produces no
duplicate buffer entry - MUST NOT write directly to the active wiki -
MUST NOT read the active wiki during ingestion --- any implementation
where TRIAGE queries existing wiki content is non-conforming

\textbf{CONTEXTUALIZE} - MUST preserve a linkout to the original
external source --- this is non-optional and cannot be traded off for
storage efficiency - MUST run in the scheduled consolidation cycle, not
at streaming ingestion time - MUST create a cold memory object for every
processed external source before producing a depth-fitted representation
- MUST NOT discard the original source after compression

\textbf{CONSOLIDATE} - MUST score buffer entries against each other
before scoring against the active wiki --- buffer-internal scoring is a
non-optional phase, not an optimization; skipping it reintroduces the
self-sealing failure mode the buffer architecture was designed to
prevent - MUST operate on a defined snapshot: a specific Git commit hash
plus a metadata index high-water mark - MUST be reproducible under a
fixed model version and runtime configuration given the same buffer
snapshot --- nondeterminism introduced by floating-point variance or
sampling is acceptable; nondeterminism introduced by different buffer
state is not - MUST produce at most one Git commit per run - MUST NOT
permanently discard minority hypotheses --- contradicting clusters
transition to branches, not deletion - MUST write all edge updates
atomically with the corresponding Git commit --- partial updates are
non-conforming

\textbf{DECAY} - MUST apply the vitality formula as specified in Section
5.3 --- local weight tuning is permitted, term elimination is not - MUST
NOT decay entries whose base gravity G\_i\^{}base remains above the
gravity-protection floor. The floor evaluates against G\_i\^{}base
rather than G\_i\^{}eff, so structural centrality alone suffices to keep
a load-bearing entry protected regardless of access recency. The floor
is implementation-defined and may be operationalized as a percentile of
the base gravity distribution, an absolute threshold, or a dynamic
function of wiki size. The required property is that the floor be
monotonic in G\_i\^{}base and comparable across entries. Entries above
the floor are protected from decay; entries below the floor are
decay-eligible. - DECAY MUST compress decay-eligible entries below the
vitality threshold rather than deleting them. Entries whose base gravity
remains above the gravity-protection floor are not decay-eligible and
therefore are not subject to vitality-driven compression.

\textbf{AUDIT} - MUST operate by temporary suspension, not permanent
deletion - MUST restore entries whose suspension demonstrably degrades
query performance - MUST NOT close minority branches without either: (a)
promotion through CONSOLIDATE when the cluster crosses the promotion
threshold, or (b) explicit archival triggered by AUDIT after confirming
the incumbent remains load-bearing and the branch has not grown across a
defined number of CONSOLIDATE cycles - MUST write all results to the
audit record before executing any state transition

\textbf{General} - The active wiki MUST remain readable during any
CONSOLIDATE or AUDIT run --- no blocking locks on the read path; this
invariant is enforced architecturally by the two-scheduler split
(Section 6.1), which separates hot-path reads from sleep-cycle writes -
Federation outbound MUST be anonymized --- personally identifying
content MUST NOT cross the device boundary; this invariant MUST be
enforced by whatever federation protocol is chosen; one approach
described in Section 6.1 is to use the proto schema itself as the
boundary contract, making identifying fields inexpressible at the
interface level - No operation MUST permanently delete any object ---
terminal states are \texttt{archived} or \texttt{expired}, never
hard-deleted; archival moves full content to cold memory (Section 4)
while preserving the index record as a tombstone, maintaining audit
history and preventing silent data loss

An implementation that satisfies all of the above invariants may still
differ substantially from others --- in how it infers working-context
depth for CONTEXTUALIZE, how it tunes vitality weights in DECAY, how it
calibrates the promotion threshold in CONSOLIDATE, and what query set it
uses for AUDIT suspension tests. These are open design decisions the
spec deliberately leaves to the implementer. The invariants define the
conformance boundary; they do not prescribe a single implementation.

Conformance tests are defined in Appendix A via worked traces. A minimal
conformance suite should verify at least the TRIAGE coherence-work
prohibition and the CONSOLIDATE buffer-internal scoring requirement, as
these are the invariants most commonly violated in naive
implementations.

\begin{center}\rule{0.5\linewidth}{0.5pt}\end{center}

\hypertarget{counterarguments}{%
\subsection{8. Counterarguments}\label{counterarguments}}

\hypertarget{this-is-a-sophisticated-echo-chamber}{%
\subsubsection{8.1 ``This is a sophisticated echo
chamber''}\label{this-is-a-sophisticated-echo-chamber}}

A single-agent companion system does reinforce its user's worldview over
time. We do not dispute this. The defenses are three. First, the
alternative --- a memory system that refuses to align with its user ---
is not actually useful as a companion; it is a general knowledge base,
which is a different product for a different purpose. Second, batched
consolidation with minority-pressure promotion is a real mechanism for
belief updating within the single-agent system, not a hope that dormant
storage will someday matter. Third, the framework does not claim to
solve echo-chamber dynamics completely at the single-agent level. It
gives three correction channels --- consolidation within the agent,
federation across agents, and base model updates from outside the
framework entirely --- and discusses each honestly.

A more specific challenge comes from THEANINE (Ong et al., arXiv
2406.10996, NAACL 2025), which explicitly discards memory removal and
instead preserves memory timelines --- memories linked by temporal and
cause-effect relations across the full history of interactions. THEANINE
is the strongest design-philosophy alternative to any pruning-based
retention, and it must be engaged directly. The companion framing leads
to a different conclusion. Operational continuity --- which memory
gravity protects --- does not require retaining all memory timelines. It
requires retaining what the user's \emph{current} work depends on. A
timeline entry describing the user's reasoning from three years ago on a
problem they no longer face has no gravity in the present wiki: nothing
currently references it, no current queries access it, and retaining it
imposes coherence overhead with no utility benefit. THEANINE's
no-removal stance avoids one failure mode (premature forgetting of
timelines that turn out to matter) at the cost of accumulation without
governance --- which is precisely the accumulation problem this
framework was designed to address. The design choice is explicit: the
companion framework accepts forgetting as a feature of operational
memory, not a defect to eliminate, and routes the responsibility for
identifying \emph{which} timelines should survive to DECAY and AUDIT
rather than to a blanket no-removal policy.

\hypertarget{act-r-already-does-consequence-weighted-retention-and-the-labs-have-shipped-the-rest}{%
\subsubsection{8.2 ``ACT-R already does consequence-weighted retention,
and the labs have shipped the
rest''}\label{act-r-already-does-consequence-weighted-retention-and-the-labs-have-shipped-the-rest}}

Full concession at the mechanism level, now broader than the original
v3.2 version of this counterargument allowed. ACT-R base-level learning
modeled activation as a function of use history since the 1990s; spaced
repetition does fitness-based retention; graph centrality predates
memory gravity. Generative Agents (Park et al., arXiv 2304.03442, UIST
2023) formalized the raw-capture-plus-scheduled-reflection pattern in
2023 with a highly cited implementation --- the
buffer-plus-consolidation architecture is prior art at the mechanism
level. Beyond cognitive architectures and academic systems, the
production labs have shipped substantially parallel mechanisms:
Anthropic reportedly ships a between-session consolidation mechanism
(referred to in community documentation as ``Auto Dream'') that performs
contradiction resolution, stale-entry pruning, and batched consolidation
--- these details are based on community reporting rather than official
Anthropic documentation. OpenAI's ChatGPT ships consequence-weighted
retention through Saved Memories and Reference Chat History. SleepGate
(arXiv 2603.14517) implements sleep micro-cycles at the KV-cache layer.
D-Mem (arXiv 2603.18631) keeps a fast retrieval path alongside a Full
Deliberation fallback that processes raw dialogue history, with a
learned Quality Gating policy between them. Second Me (arXiv 2503.08102)
builds a working single-user persistent memory system. At the mechanism
level, this paper is not proposing anything that a well-read reviewer
will consider new.

The contribution is not the retention formula, nor the sleep-modeled
consolidation architecture, nor minority-hypothesis retention, nor
gravity protection, nor the general idea of context-sensitive
compression for a specific reader. All of these have parallels in
published or deployed work. The contribution is: (1) the normative
system-class definition --- not just the word ``companion'' but the
specification of evaluation target and retention-policy obligations that
follow from treating a system as one; (2) the time-structured procedural
conflict rule for the mirror-vs-compensate tension, operationalized
across streaming/consolidation/audit timescales; (3) effective
minority-hypothesis influence as a falsifiable prediction targeting a
structural failure mode no existing benchmark captures; and (4)
architectural separability as a prerequisite for base-model evolution to
function as an external correction channel. These are framing
contributions that give the mechanisms meaning in a specific design
context. The contribution is upstream of the mechanisms the field is
shipping.

This is a defensible position but only if the paper is honest about the
scope of the concession. Nothing here invalidates the framing
contributions in Section 1 and the design principle in Section 1.2.
Everything here sharpens what ``framing contribution'' actually means:
it means the paper is useful as vocabulary for an architectural space
that has been built into but not publicly theorized in this form.

\hypertarget{companion-systems-are-dangerous-because-they-reinforce-bad-beliefs}{%
\subsubsection{8.3 ``Companion systems are dangerous because they
reinforce bad
beliefs''}\label{companion-systems-are-dangerous-because-they-reinforce-bad-beliefs}}

This is the hardest counterargument and it deserves the most space. A
memory system explicitly designed to align with its user will reinforce
whatever the user already believes, including false, harmful, or
delusional beliefs. Naming the class honestly is a precondition for
safety but it is not a safety mechanism by itself.

The framework's response has four layers.

\textbf{Layer 1: Honest naming.} The risk is acknowledged as a
structural property of the system class. Pretending a personal AI memory
system is a neutral truth-tracker when it is actually drifting with its
user hides the failure mode under a claim of objectivity, and users
trust such systems more precisely when they should trust them less.
Naming is a precondition.

\textbf{Layer 2: Compensate mechanisms.} AUDIT and minority-hypothesis
retention are specifically designed to resist entrenchment and
monoculture. They are limited --- AUDIT sensitivity is an open problem
and minority retention alone is museum storage --- but they are real
mechanisms doing real work on the compensate side of the
mirror-vs-compensate principle.

\textbf{Layer 3: Batched consolidation.} The sleep-function architecture
is the framework's central compensate mechanism. Minority-pressure
promotion gives accumulated contradictory evidence a structural path to
shifting the dominant interpretation during a consolidation window. This
is not a hope that stored alternatives will be resurfaced; it is a
mechanism for integrating them when buffer pressure crosses a threshold.
The framework can update beliefs through this channel without claiming
to be a truth-tracker --- the update is pragmatic, driven by accumulated
pressure from the user's own subsequent experience, not by external
correspondence.

\textbf{Layer 4: Architectural separability preserves an external
correction channel.} The separation of external memory from model
weights is an established architectural doctrine, not a new claim. Lewis
et al.~(2020) already justifies it on operational grounds: external
knowledge can be ``revised and expanded'' without retraining. The Atlas
architecture (Izacard et al., arXiv:2208.03299) extends this: knowledge
stores can be ``kept up-to-date without retraining, by updating or
swapping their index at test time.'' The harness engineering review
({[}53{]}) goes further, explicitly recommending cross-model transfer
tests as standard practice and arguing that reliability gains in memory
systems come from modifying the environment around the base model rather
than the model itself. Within this established externalization doctrine,
what this paper adds is a \emph{companion-specific safety rationale}
that prior work does not articulate: separability is not merely
operationally convenient (updates without retraining, provenance
auditing) but structurally necessary for base-model evolution to
function as an external correction channel specifically against
user-coupled epistemic entrenchment. That rationale is narrower than
``keep knowledge external.'' It is a safety commitment with a named
mechanism: a user running a companion system for five years benefits
from the model's improved factual priors and alignment training
precisely because swapping the base model is a configuration change, not
a wiki operation. Fold the wiki into weights and this channel closes
permanently.

This is a real correction channel but an incomplete one. Three honest
limits: the wiki still anchors interpretation, so high-gravity false
entries still bias outputs; base model updates are not always
corrections, since labs update for many reasons including some that may
make the system worse for a specific user; and the user does not control
when updates happen. The framework benefits from this channel without
being able to rely on it. What the framework \emph{does} contribute is
architectural restraint --- the commitment not to fold the companion
layer into weights, which would eliminate this channel entirely.
Separability is a design commitment, not an implementation detail.

Taken together, the four layers constitute an incomplete but non-trivial
safety story. They do not solve the reinforcement-of-bad-beliefs
problem. They give it structural defenses on three different timescales
(scheduled within-agent consolidation cycles, cross-agent federation,
base-model evolution across model generations) and name the remaining
gap as open work.

\begin{center}\rule{0.5\linewidth}{0.5pt}\end{center}

\hypertarget{limitations}{%
\subsection{9. Limitations}\label{limitations}}

\begin{itemize}
\tightlist
\item
  \textbf{AUDIT sensitivity is the critical open problem.} The
  framework's compensate mechanisms ultimately depend on AUDIT being
  able to detect when a high-gravity entry is producing bad outcomes. If
  the query set used for stress testing is narrow or self-confirming,
  entrenchment survives. This is where follow-up work should
  concentrate.
\item
  \textbf{Architectural commitment: do not fold the wiki into weights.}
  The external correction channel in Section 8.3 depends on the
  companion layer remaining separable from the base model weights. Model
  editing research may eventually make weight-level integration
  technically possible; the framework's safety story specifically rules
  this out as a design move. This is a principled constraint, not a
  current limitation to overcome.
\item
  \textbf{Operational definition of \texttt{task\_predictive\_utility}.}
  Credit assignment under delayed and noisy user feedback is a real
  problem the framework does not solve. The utility signal is the
  channel through which pragmatic truth concerns re-enter (Section 2.4),
  but its current operationalization is shallow.
\item
  \textbf{Working-context depth inference is an open modeling problem.}
  CONTEXTUALIZE infers the user's depth of engagement from query
  patterns and wiki topology rather than requiring explicit
  specification. If that inference is wrong, the operation produces
  systematic compression distortion that propagates into the active wiki
  during the next CONSOLIDATE cycle. Linkout preservation is a partial
  mitigation --- the original source remains reachable --- but anchoring
  effects mean users may rely on the compressed entry without inspecting
  the source. Reliable inference of working-context depth from
  behavioral signals is an open research problem this paper does not
  resolve.
\item
  \textbf{CONTEXTUALIZE overlaps conceptually with personalized
  summarization and adaptive hypermedia.} The idea of compressing
  content differently depending on the reader's role or task has
  established precedent in adaptive hypermedia {[}3{]} and automatic
  summarization (Nenkova \& McKeown, 2011; Mani, 2001). The framework's
  contribution is the architectural placement of this logic in the dream
  cycle, its coordination with CONSOLIDATE and TRIAGE, and its
  application at the retention-policy level for a companion-memory
  class. The concept is not novel; the coordination bundle is.
\item
  \textbf{Cost model gap.} The metabolic layer has its own overhead.
  TRIAGE is cheap, DECAY is continuous, CONTEXTUALIZE and CONSOLIDATE
  are batched and potentially expensive, and AUDIT is occasional. We
  have not characterized the combined cost against the coherence,
  fragility-resistance, and variance-maintenance benefits the framework
  aims to provide.
\item
  \textbf{No user evidence for the class.} The taxonomic claim in
  Section 1.1 is a classification move supported by analogy to adjacent
  domains (HCI, PKM, user modeling). We do not present user studies
  showing that personal LLM memory users actually prefer coherence over
  correction. That is a first-order empirical open problem.
\item
  \textbf{No current safety mechanism for fully novel bad beliefs.} If a
  user arrives at a bad belief not represented anywhere in the base
  model and not contradicted by subsequent experience, none of the four
  layers in Section 8.3 catches it. This is the residue after the safety
  story lands, and we do not currently know what a principled mechanism
  for it looks like in a companion system.
\end{itemize}

The circularity of coherence measurement is not a limitation (Section
1.3). It is part of the thesis.

\begin{center}\rule{0.5\linewidth}{0.5pt}\end{center}

\hypertarget{research-agenda-three-correction-channels}{%
\subsection{10. Research Agenda: Three Correction
Channels}\label{research-agenda-three-correction-channels}}

The following directions are proposed as a forward-looking research
agenda, not as a summary of existing work; the federation unit types ---
family, team, department, community --- are design proposals for future
investigation, not descriptions of deployed systems.

We name three correction channels operating on different timescales.
Each opens a distinct research direction.

\hypertarget{within-agent-consolidation-cycles-and-minority-pressure-promotion}{%
\subsubsection{10.1 Within-agent: consolidation cycles and
minority-pressure
promotion}\label{within-agent-consolidation-cycles-and-minority-pressure-promotion}}

The single-agent consolidation mechanism described in Section 5.5 is the
fastest correction channel --- it runs on a schedule measured in hours
or days. Open work: what scoring function weights buffer-internal
agreement against wiki agreement? How should minority-pressure
thresholds be tuned? How does consolidation interact with AUDIT cycles
when they overlap? What does the consolidation window look like for use
cases with sparse ingestion (a casual personal knowledge base) versus
dense ingestion (a professional research system)?

\hypertarget{cross-agent-federation-across-named-unit-types}{%
\subsubsection{10.2 Cross-agent: federation across named unit
types}\label{cross-agent-federation-across-named-unit-types}}

Federation is a distinct research layer, not a rescue for single-agent
problems. The relevant units are not ``multiple companion systems'' in
the abstract. They are specific, structurally different organizational
forms, each with its own update dynamics.

\textbf{Family wikis} have the longest time horizons and genuine
generational turnover. Planck's mechanism applies literally --- new
members are born into an existing worldview, inherit parts of it, and
bring new context that the federation must integrate. Open problems
include how inherited gravity transfers across generations and which
entries stabilize across three generations versus one.

\textbf{Team and company wikis} update at the pace of hiring and
offboarding. Consolidation cycles must handle the integration of new
members as structurally different from the integration of new
information --- a new team member arrives with their own prior worldview
that must partially mirror the team's and partially update it.

\textbf{Department wikis} sit between team and organizational scales.
Role handoffs are slower than team churn but faster than generational
turnover. Explicit authority structures affect what counts as a
high-gravity entry, which means the framework's gravity model must
accommodate formal authority as a signal distinct from organic
dependency count.

\textbf{Community wikis} --- open-source projects, research communities,
interest groups --- have self-selecting membership and the most complex
drift dynamics because there is no formal boundary around federation
membership. Existing literature on communities of practice and
open-source knowledge management is directly relevant here and we do not
attempt to summarize it.

Open research problems across all four unit types:
disagreement-resolution mechanisms that preserve coherence for both
sides, anonymized sync protocols that share gravity and minority signal
without exposing content, and turnover dynamics at the federation level.

Federated companion memory could also provide a richer source of
grounded human judgment than current post-training signals: not just
pairwise preferences, but consolidation decisions made by real users
integrating real knowledge over time, with the surrounding context of
what they audited away and what minority hypotheses eventually won. We
name this as a research direction without committing to any particular
training methodology.

\hypertarget{external-base-model-evolution-preserved-by-separability}{%
\subsubsection{10.3 External: base model evolution preserved by
separability}\label{external-base-model-evolution-preserved-by-separability}}

The slowest correction channel operates on the timescale of base model
generations --- months to years. It is not implemented by the framework;
it is \emph{preserved} by the framework's architectural commitment to
separability. The research direction is less about building new
mechanisms and more about characterizing how companion systems should
respond to base model swaps. Should consolidation cycles be re-run after
a base model update? How should gravity scores adjust when the
underlying reasoning engine has different factual priors? What does
version compatibility look like between a wiki seeded against one base
model and a subsequent generation?

These three channels operate on different timescales (scheduled
consolidation cycles measured in hours to days, cross-agent federation
measured in weeks to years, base-model evolution measured in months to
years), address different failure modes, and compose without
conflicting. No single channel is sufficient. The framework's safety
story is the combination.

\hypertarget{application-context-ai-assisted-software-development}{%
\subsubsection{10.4 Application Context: AI-Assisted Software
Development}\label{application-context-ai-assisted-software-development}}

The three preceding subsections describe research directions within the
governance framework itself. A fourth direction concerns where the
framework applies.

One domain where the failure modes addressed in this paper become
particularly acute is AI-assisted software development. Long-lived
software systems accumulate interdependent artifacts --- including code,
architectural decisions, requirements, and operational signals --- that
must remain coherent over extended time horizons. In current AI-assisted
tooling, this continuity is maintained implicitly by human actors rather
than by the tools themselves.

In such environments, the absence of explicit retention-governance
mechanisms leads to predictable degradation: architectural intent is
lost, contradictory assumptions accumulate, and previously resolved
decisions are repeatedly revisited. The companion-memory failure modes
described in this paper --- entrenchment under user-coupled drift,
suppression of minority evidence, and coherence-preserving fragmentation
--- therefore compound at system scale rather than remaining localized.

Whether the framework is sufficient for such multi-artifact and
potentially multi-agent environments, and how it interacts with
domain-specific mechanisms for engineering knowledge, remains open ---
particularly in settings where memory is distributed across tools,
teams, and evolving system boundaries.

\begin{center}\rule{0.5\linewidth}{0.5pt}\end{center}

\hypertarget{conclusion}{%
\subsection{11. Conclusion}\label{conclusion}}

This paper proposes a retention-and-governance profile for single-user
companion knowledge systems built on the LLM wiki pattern: a set of
normative obligations, an object model, conformance invariants, and a
time-structured procedural rule addressing the specific failure mode of
entrenchment under user-coupled drift. Positioned as a
companion-specific specialization within the 2026 landscape of emerging
governance frameworks --- including Context Cartography ({[}45{]}) and
MemOS ({[}24{]}) --- its sharpest contribution is the multi-cycle buffer
pressure accumulation mechanism that gives accumulated minority evidence
a structural path to influence centrality-protected dominant
interpretations, a structural failure mode no existing operator or
governance framework explicitly targets.

Personal LLM memory systems already exhibit user-coupled retention
dynamics that can accumulate into drift over time. This is sufficient
grounds to propose normative governance obligations for treating such
systems as a distinct design class --- a class that prior work has named
(MemoryBank uses ``companion scenario''; Second Me builds explicitly for
one user; LongMemEval ties evaluation to personalized assistance
contexts) but not governed with a normative retention policy. The
governance profile this paper provides implements the
mirror-vs-compensate principle with a raw buffer, streaming TRIAGE,
continuous DECAY, dream-cycle CONTEXTUALIZE for depth-fitted compression
of external sources, batched CONSOLIDATE, and periodic AUDIT, supported
by memory gravity and minority-hypothesis retention.

The contribution is the framing and the design principle, not the
mechanisms. Most mechanisms are inherited from cognitive architectures,
recommender-system diversity injection, graph centrality, and pragmatist
epistemology, and many of them have direct parallels in production
systems shipped by Anthropic, OpenAI, and the academic lineage running
from Generative Agents (2023) through MemGPT, Mem0, SleepGate, D-Mem,
and Second Me. The reframe puts them to work on a specific design target
that, to our knowledge, current LLM memory literature has not made
explicit in this form.

Four contributions are the ones we are most confident survive engagement
with the current landscape as standalone framing contributions.

\emph{Companion system} as a normatively defined design class --- not a
new term (MemoryBank uses ``companion scenario''; Second Me builds
explicitly for one user; LongMemEval ties evaluation to personalized
assistance contexts), but a normative governance profile specifying the
retention obligations that follow from the class: the system is
evaluated on longitudinal user utility under drift, subject to explicit
anti-entrenchment and correction obligations, rather than
correspondence-based truth-tracking; it must mirror operational
continuity, it must compensate epistemic failure, and it must preserve
architectural separability to keep the base-model correction channel
open. The class is already named; what is missing is the governance
profile --- what MUST be mirrored, what MUST be compensated, and why ---
and that is what this paper provides.

The \emph{mirror-vs-compensate principle} as a time-structured
procedural conflict rule --- not just the observation that mirroring and
compensating are in tension (prior work including ``To Mask or to
Mirror,'' arXiv 2510.01924, names that tension and uses the vocabulary
explicitly), but a specific resolution procedure operationalized across
concrete architectural timescales: mirror by default in the streaming
path, compensate during scheduled consolidation windows, AUDIT as the
slow-cycle tiebreaker, and instantiated case-by-case in the conflict
routing matrix (§5.0). The specificity is the contribution.

\emph{Effective minority-hypothesis influence} as a falsifiable
prediction targeting a structural failure mode that existing benchmarks
do not capture. LongMemEval's knowledge-update category tests explicit
profile corrections. TeaFarm (THEANINE, arXiv:2406.10996) tests whether
memory changes output via counterfactual presence/absence. Prediction 4
targets something structurally different: belief revision through
multi-cycle buffer pressure accumulation under a centrality-protected
incumbent, where no single contradictory entry achieves integration
alone but a cluster achieves it together. Retention alone is not
variance. Surfacing alone is not variance. Only measurable change in
downstream outputs is effective variance, and that is the prediction the
framework is prepared to be falsified on.

\emph{Architectural separability from base-model weights} as a safety
design commitment whose specific rationale has not been articulated in
prior work: within the externalization doctrine established by Lewis et
al.~(2020) and Atlas (Izacard et al., arXiv:2208.03299), separability is
not just operationally convenient but structurally necessary for
base-model evolution to function as an external correction channel
against user-coupled epistemic entrenchment. A user running a companion
system for five years benefits from the model's improved factual priors
and alignment training \emph{for free} because swapping the base model
is a configuration change, not a wiki operation. Fold the wiki into
weights and this channel closes permanently.

Two further moves are important to the coherence of the framework but
are stated more cautiously. First, \emph{circularity-as-thesis} is the
governing stance that makes companion-memory coherence intentional
rather than an unacknowledged defect --- AUDIT operationalizes a Kuhnian
discipline by exposing accumulated anomaly debt at the entry level, as
grounded by the discussion in Section 5.8, rather than letting it
accumulate invisibly at the wiki level, making the circularity
manageable rather than eliminating it. Second, \emph{CONTEXTUALIZE}
extends the metabolic framing from retention to ingestion by treating
depth-fitted compression as selective absorption during the dream cycle
--- coordinating depth inference, compression target, linkout
preservation, and scheduling as a bundle that prior work (Generative
Agents, D-Mem, LightMem) does not assemble together.

The honest framing: the labs and the academic literature have built much
of the machinery, and what this paper offers is design vocabulary for
evaluating and combining that machinery in a specific class of system.
That is a narrower claim than ``the field lacks the language,'' and it
is the one we mean.

The framework does not self-correct on epistemic grounds at the
single-agent level. It does give structural defenses on three timescales
--- within-agent consolidation, cross-agent federation across family,
team, department, and community wikis, and external base model updates
preserved by architectural separability. None individually solves the
reinforcement-of-bad-beliefs problem. Together they make the safety
story non-trivial.

Most systems today do not state clearly what kind of memory system they
are building. That is the gap this paper closes. The safety story is
partial mitigation, not solved architecture --- structural defenses on
three timescales against a failure mode that persists. What remains to
be built is on the compensate side, and we have tried to name the
specific directions where the work is.

\begin{center}\rule{0.5\linewidth}{0.5pt}\end{center}

\hypertarget{references}{%
\subsection{References}\label{references}}

{[}1{]} Anderson, J. R., \& Lebiere, C. (1998). \emph{The Atomic
Components of Thought}. Lawrence Erlbaum Associates.

{[}2{]} Bonawitz, K., et al.~(2019). Towards federated learning at
scale: A system design. In \emph{Proceedings of MLSys 2019}.
arXiv:1902.01046.

{[}3{]} Brusilovsky, P. (2001). Adaptive hypermedia. \emph{User Modeling
and User-Adapted Interaction}, 11, 87--110.

{[}4{]} Chhikara, P., Khant, D., Aryan, S., Singh, T., \& Yadav, D.
(2025). Mem0: Building production-ready AI agents with scalable
long-term memory. arXiv:2504.19413.

{[}5{]} Dewey, J. (1938). \emph{Logic: The Theory of Inquiry}. Henry
Holt and Company.

{[}6{]} Doyle, J. (1979). A truth maintenance system. \emph{Artificial
Intelligence}, 12(3), 231--272.

{[}7{]} Ebbinghaus, H. (1885). \emph{Über das Gedächtnis}. Duncker \&
Humblot.

{[}8{]} Fang, J., Deng, X., Xu, H., Jiang, Z., Tang, Y., Xu, Z., Deng,
S., Yao, Y., Wang, M., Qiao, S., Chen, H., \& Zhang, N. (2026).
LightMem: Lightweight and efficient memory-augmented generation.
\emph{ICLR 2026}. arXiv:2510.18866.

{[}9{]} Ford, N., Parsons, R., \& Kua, P. (2017). \emph{Building
Evolutionary Architectures: Support Constant Change}. O'Reilly Media.

{[}10{]} Gärdenfors, P., \& Makinson, D. (1988). Revisions of knowledge
systems using epistemic entrenchment. In \emph{Proceedings TARK '88},
83--95.

{[}11{]} Goel, R. (2026). \emph{LLM Wiki v2} {[}GitHub gist{]}.
https://gist.github.com/rohitg00/2067ab416f7bbe447c1977edaaa681e2

{[}12{]} Hirsch, J. E. (2005). An index to quantify an individual's
scientific research output. \emph{PNAS}, 102(46), 16569--16572.

{[}13{]} Hu, Y., Liu, S., Yue, Y., Zhang, G., et al.~(2025). Memory in
the Age of AI Agents. arXiv:2512.13564.

{[}14{]} Izacard, G., Lewis, P., Lomeli, M., Hosseini, L., Petroni, F.,
Schick, T., Dwivedi-Yu, J., Joulin, A., Riedel, S., \& Grave, E. (2022).
Few-shot learning with retrieval augmented language models.
arXiv:2208.03299.

{[}15{]} James, W. (1907). \emph{Pragmatism: A New Name for Some Old
Ways of Thinking}. Longmans, Green, and Co.

{[}16{]} Jia, Z., Li, J., Kang, Y., Wang, Y., Wu, T., Wang, Q., Wang,
X., Zhang, S., Shen, J., Li, Q., Qi, S., Liang, Y., He, D., Zheng, Z.,
\& Zhu, S.-C. (2025). The AI Hippocampus: How far are we from human
memory? \emph{TMLR}. arXiv:2601.09113.

{[}17{]} Jovovich, M., \& Sigman, B. (2026). \emph{MemPalace v3.0.0}
{[}GitHub repository{]}.
https://github.com/milla-jovovich/mempalace/releases/tag/v3.0.0

{[}18{]} Ju, H., Zhou, D., Blevins, A. S., Lydon-Staley, D. M., Kaplan,
J., Tuma, J. R., \& Bassett, D. S. (2020). The network structure of
scientific revolutions. arXiv:2010.08381.

{[}19{]} Ju, H., Zhou, D., Blevins, A. S., Lydon-Staley, D. M., Kaplan,
J., Tuma, J. R., \& Bassett, D. S. (2022). Historical growth of concept
networks in Wikipedia. \emph{Collective Intelligence}, 1(2).

{[}20{]} Karpathy, A. (2026). \emph{LLM Wiki: A pattern for building
personal knowledge bases using LLMs} {[}GitHub gist{]}.
https://gist.github.com/karpathy/442a6bf555914893e9891c11519de94f

{[}21{]} Kuhn, T. S. (1962). \emph{The Structure of Scientific
Revolutions}. University of Chicago Press.

{[}22{]} Lee, I. Y., Yang, C., \& Berg-Kirkpatrick, T. (2025). Optical
Context Compression Is Just (Bad) Autoencoding. arXiv:2512.03643.

{[}23{]} Lewis, P., Perez, E., Piktus, A., Petroni, F., Karpukhin, V.,
Goyal, N., Küttler, H., Lewis, M., Yih, W., Rocktäschel, T., Riedel, S.,
\& Kiela, D. (2020). Retrieval-augmented generation for
knowledge-intensive NLP tasks. \emph{NeurIPS}, 33, 9459--9474.

{[}24{]} Li, Z., Xi, C., Li, C., Chen, D., Chen, B., Song, S., Niu, S.,
Wang, H., et al.~(2025). MemOS: A Memory OS for AI System.
arXiv:2507.03724.

{[}25{]} Liu, F., \& Qiu, H. (2025). Context Cascade Compression:
Exploring the Upper Limits of Text Compression. arXiv:2511.15244.

{[}26{]} Mani, I. (2001). \emph{Automatic Summarization}. John
Benjamins.

{[}27{]} McClelland, J. L., McNaughton, B. L., \& O'Reilly, R. C.
(1995). Why there are complementary learning systems in the hippocampus
and neocortex. \emph{Psychological Review}, 102(3), 419--457.

{[}28{]} Men, X., Xu, M., Zhang, Q., Wang, B., Lin, H., Lu, Y., Han, X.,
\& Chen, W. (2024). ShortGPT: Layers in large language models are more
redundant than you expect. arXiv:2403.03853.

{[}29{]} Nenkova, A., \& McKeown, K. (2011). Automatic summarization.
\emph{Foundations and Trends in Information Retrieval}, 5(2--3),
103--233.

{[}30{]} Ong, K. T., Kim, N., Gwak, M., Chae, H., Kwon, T., Jo, Y.,
Hwang, S., Lee, D., \& Yeo, J. (2025). Towards lifelong dialogue agents
via timeline-based memory management. In \emph{Proceedings of NAACL
2025}. arXiv:2406.10996.

{[}31{]} Packer, C., Wooders, S., Lin, K., Fang, V., Patil, S. G.,
Stoica, I., \& Gonzalez, J. E. (2023). MemGPT: Towards LLMs as operating
systems. arXiv:2310.08560.

{[}32{]} Page, L., Brin, S., Motwani, R., \& Winograd, T. (1999).
\emph{The PageRank citation ranking: Bringing order to the web}.
Stanford InfoLab.

{[}33{]} Park, J. S., O'Brien, J. C., Cai, C. J., Morris, M. R., Liang,
P., \& Bernstein, M. S. (2023). Generative agents: Interactive simulacra
of human behavior. In \emph{Proceedings of UIST 2023}. arXiv:2304.03442.

{[}34{]} Peirce, C. S. (1878). How to make our ideas clear.
\emph{Popular Science Monthly}, 12, 286--302.

{[}35{]} Planck, M. (1950). \emph{Scientific Autobiography and Other
Papers}. Williams \& Norgate.

{[}36{]} Qian, C., Parisi, A., Bouleau, C., Tsai, V., Lebreton, M., \&
Dixon, L. (2025). To mask or to mirror: Human-AI alignment in collective
reasoning. In \emph{Proceedings of EMNLP 2025}. arXiv:2510.01924.

{[}37{]} Shi, W., Gao, M., Xu, Z., Feng, S., Xu, W., Shi, P.,
Zettlemoyer, L., \& Tsvetkov, Y. (2024). LongMemEval: Benchmarking chat
assistants on long-term interactive memory. arXiv:2410.10813.

{[}38{]} Khemani, S. (2025). \emph{Reverse-engineering ChatGPT's memory
architecture} {[}community analysis; not official OpenAI
documentation{]}.
https://www.shloked.com/writing/chatgpt-memory-bitter-lesson (archived:
https://web.archive.org/web/20260413152757/https://www.shloked.com/writing/chatgpt-memory-bitter-lesson)

{[}39{]} Tononi, G., \& Cirelli, C. (2014). Sleep and the price of
plasticity. \emph{Neuron}, 81(1), 12--34.

{[}40{]} Tulving, E. (1972). Episodic and semantic memory. In
\emph{Organization of Memory}, Academic Press.

{[}41{]} Wang, S., Yu, E., Love, O., Zhang, T., Wong, T., Scargall, S.,
\& Fan, C. (2026). MemMachine: A Ground-Truth-Preserving Memory System
for Personalized AI Agents. arXiv:2604.04853.

{[}42{]} Wei, H., Sun, Y., \& Li, Y. (2025). DeepSeek-OCR: Contexts
Optical Compression. arXiv:2510.18234.

{[}43{]} Wei, J., Ying, X., Gao, T., Bao, F., Tao, F., \& Shang, J.
(2025). AI-native memory 2.0: Second Me. arXiv:2503.08102.

{[}44{]} Wu, Y., Liang, S., Zhang, C., Wang, Y., Zhang, Y., Guo, H.,
Tang, R., \& Liu, Y. (2025). From Human Memory to AI Memory: A Survey on
Memory Mechanisms in the Era of LLMs. arXiv:2504.15965.

{[}45{]} Wu, Z., \& Gartner, G. (2026). Context Cartography: Toward
Structured Governance of Contextual Space in Large Language Model
Systems. arXiv:2603.20578.

{[}46{]} Xie, Y. (2026). Learning to forget: Sleep-inspired memory
consolidation for resolving proactive interference in large language
models. arXiv:2603.14517.

{[}47{]} Xu, W., Liang, Z., Mei, K., Gao, H., Tan, J., \& Zhang, Y.
(2025). A-MEM: Agentic Memory for LLM Agents. arXiv:2502.12110.

{[}48{]} You, Z., Yuan, J., \& Cai, J. (2026). D-Mem: A dual-process
memory system for LLM agents. arXiv:2603.18631.

{[}49{]} Zadeh, L. A. (1965). Fuzzy sets. \emph{Information and
Control}, 8(3), 338--353.

{[}50{]} Zep AI. (2025). Zep: A temporal knowledge graph architecture
for agent memory. arXiv:2501.13956.

{[}51{]} Zhang, Z., Bo, X., Ma, C., Li, R., Chen, X., Dai, Q., Zhu, J.,
Dong, Z., \& Wen, J.-R. (2024). A Survey on the Memory Mechanism of
Large Language Model based Agents. arXiv:2404.13501.

{[}52{]} Zhong, W., Guo, L., Gao, Q., Ye, H., \& Wang, Y. (2023).
MemoryBank: Enhancing large language models with long-term memory.
arXiv:2305.10250.

{[}53{]} Zhou, C., Chai, H., Chen, W., Guo, Z., Shan, R., Song, Y., Xu,
T., Yang, Y., Yu, A., Zhang, W., Zheng, C., Zhu, J., Zheng, Z., Zhang,
Z., Lou, X., Zhang, C., Fu, Z., Wang, J., Liu, W., Lin, J., \& Zhang, W.
(2026). Externalization in LLM Agents: A Unified Review of Memory,
Skills, Protocols and Harness Engineering. arXiv:2604.08224.

\begin{center}\rule{0.5\linewidth}{0.5pt}\end{center}

\hypertarget{acknowledgments}{%
\subsection{Acknowledgments}\label{acknowledgments}}

This paper benefited from AI-assisted brainstorming, drafting support,
and critique during development. All design decisions, interpretations,
and claims are the author's responsibility. The author thanks the
open-source contributors behind Karpathy's LLM Wiki and MemPalace.

\begin{center}\rule{0.5\linewidth}{0.5pt}\end{center}

\hypertarget{appendix-a-worked-traces}{%
\subsection{Appendix A: Worked Traces}\label{appendix-a-worked-traces}}

These traces demonstrate legal and illegal system behavior against the
conformance invariants in Section 7.5. Their purpose is not to
illustrate the mechanism --- Section 5 does that --- but to define the
boundary between correct and incorrect use of the operation interfaces.
This is the role MapReduce's word-count example plays: it is not there
to teach word-count, it is there to show what a valid job looks like and
what the runtime guarantees about it.

\hypertarget{trace-1-legal-minority-hypothesis-promotion-across-two-consolidation-cycles}{%
\subsubsection{Trace 1 --- Legal: minority hypothesis promotion across
two consolidation
cycles}\label{trace-1-legal-minority-hypothesis-promotion-across-two-consolidation-cycles}}

\textbf{Setup.} Active wiki contains entry X with the
\texttt{gravity-protected} flag set --- an incumbent belief with high
dependency count. Over two weeks, five buffer entries arrive that
individually contradict X but mutually support each other.

\textbf{TRIAGE (entries 1--5, streaming).} Each entry passes the shallow
filter. Each is assigned a content hash ID, ingestion timestamp, and
origin channel. No coherence work is performed --- TRIAGE does not read
the active wiki. All five enter the buffer in \texttt{pending} state.
TRIAGE did not write to the active wiki at any point during this phase
--- all entries entered the system exclusively through the buffer. This
is conforming: TRIAGE fulfilled its obligation and no more.

\textbf{CONSOLIDATE cycle 1.} Buffer-internal scoring shows three of the
five entries mutually supporting each other. Wiki scoring shows each
contradicts X individually --- none crosses the promotion threshold
alone. CONSOLIDATE does not integrate any entry into the active wiki.
Instead, it creates minority branch B1 referencing incumbent X, writes
entries 1--3 to B1 in \texttt{open} state, and leaves entries 4--5 in
extended buffer. One Git commit records the branch creation. One DB
transaction updates the edges index. \emph{Conforming: CONSOLIDATE
operated on a defined snapshot, produced one commit, and did not discard
the minority cluster.}

\textbf{CONSOLIDATE cycle 2.} Two additional supporting entries arrived
since cycle 1. Buffer-internal scoring now shows five
mutually-supporting entries. Combined pressure crosses the promotion
threshold. CONSOLIDATE produces one Git commit: X transitions to
\texttt{decaying}, branch B1 is merged to main, new consolidated entry Y
enters \texttt{active} state. One DB transaction updates entries and
edges atomically. \emph{Conforming: deterministic given the snapshot,
one commit, no minority branch deleted.}

\textbf{AUDIT (next scheduled cycle).} X is in \texttt{decaying} state
--- not removed. AUDIT runs suspension tests against X's remaining
references. Query performance is unchanged. X gravity is reduced. X
eventually reaches \texttt{archived} state through DECAY. Audit record
created before any state transition. \emph{Conforming.}

\textbf{Result: legal.} Minority position achieved integration through
multi-cycle buffer pressure accumulation. All invariants satisfied. No
entry was permanently deleted. TRIAGE performed no coherence work.

\hypertarget{trace-2-illegal-triage-performing-coherence-work}{%
\subsubsection{Trace 2 --- Illegal: TRIAGE performing coherence
work}\label{trace-2-illegal-triage-performing-coherence-work}}

\textbf{Setup.} Active wiki contains entry X with the
\texttt{gravity-protected} flag set. A new buffer entry E arrives that
directly contradicts X.

\textbf{Illegal behavior.} TRIAGE queries the active wiki, detects the
contradiction, and routes E directly to quarantine. The buffer log shows
E as \texttt{rejected} at ingestion time with reason ``contradicts
high-gravity entry X.''

\textbf{Why this is a violation.} TRIAGE MUST NOT read the active wiki
during ingestion. TRIAGE MUST NOT perform semantic contradiction
resolution. Both invariants are violated. The consequence is the
self-sealing failure mode the buffer architecture was designed to
prevent: if TRIAGE quarantines every entry that contradicts a
gravity-protected incumbent, the dominant interpretation never receives
minority pressure and cannot update through the consolidation mechanism
regardless of how much contradicting evidence accumulates. Critically,
by routing E to quarantine rather than the buffer, this illegal behavior
also bypasses the minority-pressure accumulation mechanism entirely ---
future supporting entries that might have joined E in the buffer to form
a promotion-eligible cluster will never find it there, permanently
foreclosing the belief-revision path that CONSOLIDATE provides.

\textbf{The conforming behavior.} E passes TRIAGE. E enters the buffer
in \texttt{pending} state with its content hash, source pointer, and
ingestion timestamp. At the next CONSOLIDATE cycle, E is scored
individually against X --- below the promotion threshold --- and is
written to a minority branch referencing X. If more supporting entries
arrive, that branch accumulates pressure across subsequent cycles.
TRIAGE's only job was the shallow filter. It fulfilled it.

\textbf{The distinguishing test for implementations.} If an
implementation's TRIAGE operation ever reads the active wiki or scores
entries against existing wiki content, it is non-conforming ---
regardless of whether its outcomes look correct in a given test case.

\begin{center}\rule{0.5\linewidth}{0.5pt}\end{center}

\emph{Preprint v3.642. April 2026.} \emph{Contact:
stefan.miteski@ext.code.berlin} \emph{This paper is deposited on Zenodo
(concept DOI 10.5281/zenodo.19501651, citing all versions); an arXiv
version is forthcoming.}

\end{document}